\documentclass[10pt,twocolumn,letterpaper]{article}

\usepackage{cvpr}
\usepackage[pagebackref=true,breaklinks=true,letterpaper=true,colorlinks,bookmarks=false]{hyperref}
\usepackage{amsmath}
\usepackage{times}
\usepackage{graphicx}
\usepackage{color}
\usepackage{array}
\usepackage{wrapfig}

\cvprfinalcopy 

\def\cvprPaperID{4783}

\title{Divergence Prior and Vessel-tree Reconstruction}

\author{
Zhongwen Zhang \hspace{5ex} Egor Chesakov \\ University of Western Ontario \\ Canada
\and
Dmitrii Marin \hspace{5ex}  Yuri Boykov\\ University of Waterloo \\ Canada 
}

\ifcvprfinal\fi

\begin{document}

\maketitle

\begin{abstract}
We propose a new geometric regularization principle for reconstructing vector fields
based on prior knowledge about their divergence. As one important example 
of this general idea, we focus on vector fields modelling blood flow pattern
that should be divergent in arteries and convergent in veins. We show that this previously ignored 
regularization constraint can significantly improve the quality of vessel tree reconstruction particularly around  
bifurcations where non-zero divergence is concentrated. Our divergence prior is critical for resolving (binary)
sign ambiguity in flow orientations produced by standard vessel filters, \eg Frangi. 
Our vessel tree centerline reconstruction combines divergence constraints  
with robust curvature regularization. Our unsupervised method can reconstruct complete
vessel trees with near-capillary details on synthetic and real 3D volumes.
\end{abstract}

\section{Background on vessel detection} 
\label{sec:intro}

There is a large body of prior work on estimation of vessels in computer vision and biomedical imaging 
communities \cite{mattos:cmpb18}. Typically, pixel-level detection of tubular structures is based on multiscale 
eigen analysis of raw intensity Hessians developed by Frangi et al. \cite{frangi1998multiscale} 
and other research groups \cite{enquobahrie2007vessel}.  At any given point (pixel/voxel) 
such vessel enhancement filters output {\em tubularness measure} and estimates of vessel's scale 
and orientation, which describes the flow direction upto to a sign. 
While such local analysis of Hessians is very useful, 
simple thresholding of points with large-enough vesselness measure is often unreliable as 
a method for computing vessel tree structure. While thresholding works well for detecting 
relatively large vessels, detection of smaller vessels is complicated by noise,
partial voluming, and outliers (e.g. ring artifacts). More importantly, standard tubular filters
exhibit signal loss at vessel bifurcations as those do not look like tubes.

Regularization methods can address vessel continuation problems due to noise, outliers, 
and signal loss at thinner parts and bifurcations. We propose a new regularization prior
based on knowledge of the flow pattern divergence. This prior is critical for disambiguating
flow directions, which provide important cues about the vessel tree structure. 
Next subsections outline existing regularization methods 
for vessel reconstruction and motivate our approach.

It may be also interesting to apply deep learning to vessel tree detection,
but neural network training is problematic since  vessel tree ground truth 
is practically impossible in real 3D data. Practical weakly-supervised training may require
regularized loss functions \cite{rloss:eccv18} appropriate for vessel tree detection. 
While our regularization methodology may help to design such losses, 
we leave this for future work.

\subsection{Vessel representation: centerline or segment} 
\label{sec:representation}

Two common approaches to representing vessels in reconstruction methods are 
volumetric binary mask and centerline.  
Volumetric mask is typical for techniques directly computing vessel segmentation, 
\ie binary labeling of pixels/voxels. In contrast, centerline is a 1D abstraction of the vessel. 
But, if combined with information about vessel radii, it is easy to obtain a volumetric mask 
or segmentation from the vessel's centerline, e.g. using MAT \cite{siddiqi2008medial}. 
Vice versa, centerline could be estimated from the vessel's binary mask using skeletonization algorithms. 

In the context of regularization methods for vessel reconstruction, centerline representation offers significant 
advantages since powerful higher-order regularizers are easier to apply to 1D structures. For example,
centerline's curvature can be regularized \cite{thin:iccv15}, while conceptually comparable regularization 
for vessel segmentation requires optimization of Gaussian or minimum curvature of the vessel's surface 
with no known practical algorithms.  In general, curvature remains a challenging regularization criteria for surfaces
\cite{shoenemann-etal-iccv-2009,strandmark2011curvature,heber-et-al-eccv-2012,olsson2013partial,curvature:cvpr14}.
Alternatively, some vessel segmentation methods use simpler first-order regularizers producing 
minimal surfaces. While tractable, such regularizers impose a wrong prior 
for surfaces of thin structures due to their bias to compact blob shapes (a.k.a. shrinking bias).

\subsection{Towards whole tree centerline} 
\label{sec:whole_tree}

Many vessel reconstruction methods directly compute centerlines of different types 
that can be informaly defined as simplified (e.g. regularized)
1D representation of the blood flow {\em pathlines}.
For example, A/B shortest path methods reqire a user to specify two end points of a vessel
and apply Dijkstra to find an optimal pathline on a graph with edge weights based on
vesselness measure. 

Interactive A/B methods are not practical for large vessel tree reconstraction problems. While it is OK to ask a user
to identify the tree {\em root}, manual identification of all the end points ({\em leaves}) is infeasible.
There are {\em tracing} techniques \cite{aylward2002initialization} designed to trace vessel tree from a given root based on vesselness measures and some local continuation heuristics. 
Our evaluations on synthetic data with groud truth show that local tracing methods do not work well for large 
trees with many thin vessels even if we use the ground truth to provide all tree leaves
as extra seeds in addition to the root.

Our goal is unsupervised reconstruction of the whole vessel tree centerline. We optimize 
a global objective function for a field of centerline tangents. Such objectives can combine 
vesselness measure as unary potentials with different regularization constraints addressing 
centerline completion. Related prior work using centerline curvature regularization 
is reviewed in the next subsection.

\subsection{Curvature regularization for centerline} 
\label{sec:curvature}

Curvature, a second-order smoothness term, is a natural regularizer for thin structures. In general,
curvature was studied for image segmentation
\cite{shoenemann-etal-iccv-2009,strandmark2011curvature,schoenemann-etal-ijcv-2012,Pock:JMIV12,heber-et-al-eccv-2012,olsson2013partial,curvature:cvpr14,thin:iccv15},
for stereo or multi-view-reconstruction 
\cite{li2010differential, olsson2013defense,woodford2009global}, connectivity measures in analysis of diffusion MRI \cite{momayyezsiahkal20133d}, 
for tubular structures extraction \cite{thin:iccv15}, 
for {\em inpainting} \cite{alvarez1992image,chan2001nontexture} and edge completion 
\cite{guy1993inferring, williams1997stochastic, alter1998extracting}.

Olsson et al.~\cite{olsson2012curvature} propose curvature approximation 
for surface fitting regularization. Their framework employs tangential approximation of surfaces.  
The authors assume that the data points are noisy readings of the surface. 
The method estimates local surface patches, which are parametrized by a tangent plane. 
It is assumed that the distance from the data point to its tangent plane is 
a surface norm. That implicitly defines the point of tangency.

\begin{figure}[t]
\centering
\includegraphics[width=0.7\linewidth]{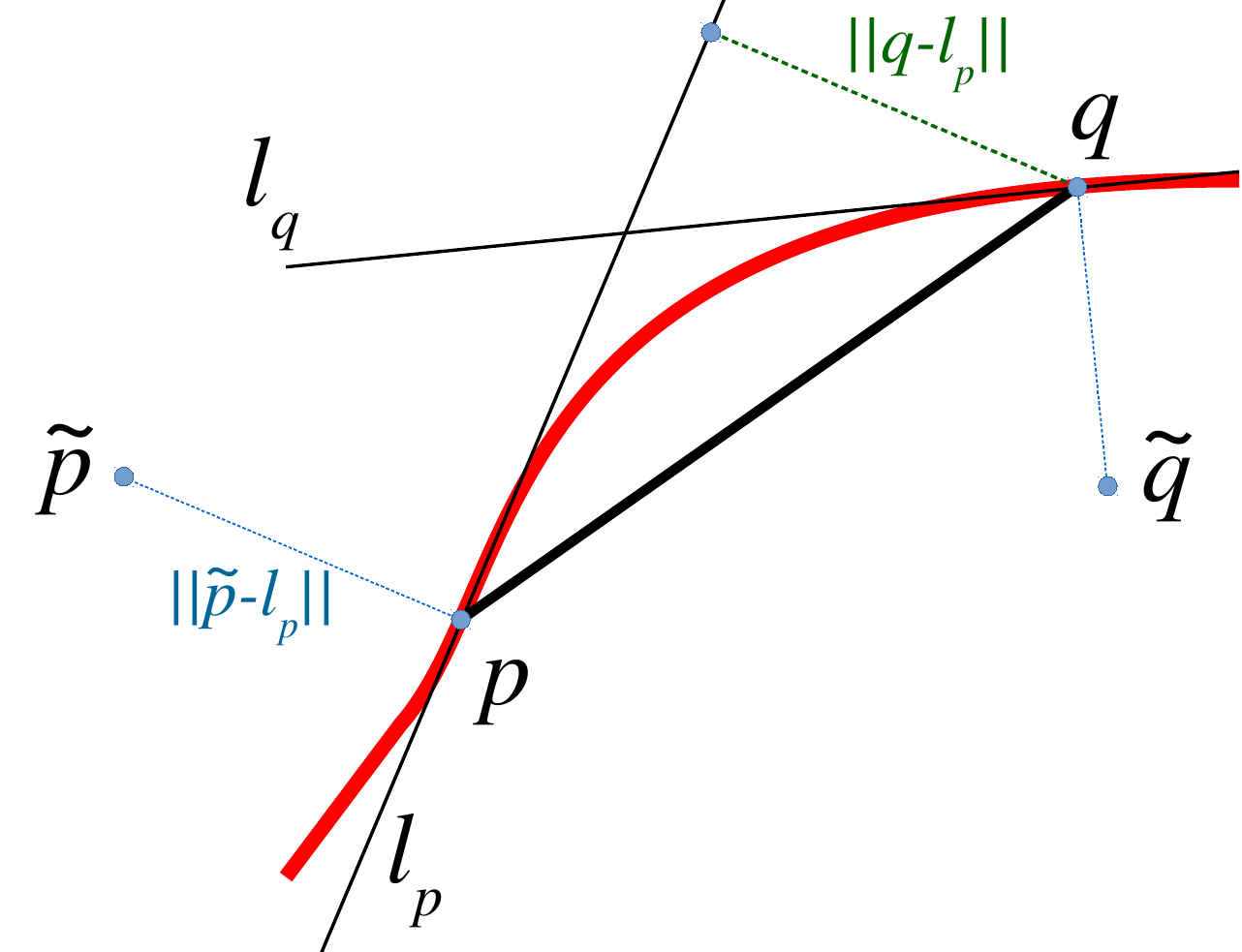} 
\caption{Curvature model of \cite{olsson2012curvature}. 
Given two points $p$ and $q$ on the red curve and
two tangents $l_p$ and $l_q$ at these points, 
the integrals of curvature are approximated by 
(\ref{eq:abs curv}--\ref{eq:sqr curv}).}
\label{fig:olsson-boykov-curv}
\end{figure}

Assume there is a smooth curve, see Fig.~\ref{fig:olsson-boykov-curv}. 
Points $p$ and $q$ on the curve and tangents $l_p$ and $l_q$ at these points are given. 
Then the integrals of curvature $\kappa(\cdot)$ is estimated by
\begin{equation}\label{eq:abs curv}
\int_p^q |\kappa(s)|ds \;\; \approx \;\; \frac{\|p - l_q\|}{\| p - q \|},
\end{equation}
\begin{equation}\label{eq:sqr curv true}
\int_p^q |\kappa(s)|^2ds \;\; \approx \;\; \frac{\|p - l_q\|^2}{\| p - q \|^3}.
\end{equation}
where $\|p - l_q\|$ is the distance between point $p$ and the tangent
line at point $q$ represented by collinear vector $l_q$.
\cite{olsson2012curvature} explores properties of these approximations 
and argues 
\begin{equation}\label{eq:sqr curv}
\kappa_{pq}(l_p,l_q) := \frac12\frac{\|p - l_q\|^2 + \|q - l_q\|^2}{\| p - q \|^2}
\end{equation}
is a better regularizer, where we used a symmetric version of integral in \eqref{eq:sqr curv true}.

Marin et al.~\cite{thin:iccv15} generalized this surface fitting problems
to detection problems where majority of the data points, 
\eg image pixels, do not belong to a thin structure. In order to do that they introduced 
binary variables in their energy indicating if a data point belongs to the thin structure. 
One of their applications is vessel detection. The proposed vessel-tree extraction system
includes vessel enhancment filtering, non-maximum suppresion for data reduction,
tangent approximation of vessels' centerline and minimum spanning tree 
for topology extraction. Assuming that detection variables are computed, 
the tangent approximation of vessels' centerline is found by minimizing energy
\begin{equation}\label{eq:unoriented curv energy}
E_u(l)=\sum_p{\| \tilde p - l_p \|^2} + \gamma \sum_{(p, q)\in N} \kappa_{pq}(l_p, l_q)
\end{equation}
where summations are over detected vessel points,
$\tilde p$ is the original data point's location, 
$l_p$ is the tangent vector at point $p$,
the denoised point location $p$ is constraint to be the closest point on tangent line at $p$, 
and $N \subset \Omega^2$ is the neighbourhood system. The curvature term in the energy 
makes the tangents ``collapse'' onto one-dimensional centerline as in
Fig.~\ref{fig:triangleArtifact}(a,c). But the same figures also show 
artifacts around bifurcations where undesired triangular structures forms 
indicating unoriented tangent model limitations.

Our experiments employs the same components as in \cite{thin:iccv15}.
Our work focuses on analysis of failure cases and improvement of the regularization stage
for tangent approximation. In particular we will show the drawbacks of 
curvature models (\ref{eq:abs curv}-\ref{eq:sqr curv}) in the context of
vessel tree extraction and propose a solution
leading to significant improvement of the results.


\subsection{Our contributions and motivation} 
\label{sec:contributions}

\begin{figure*}
\centering
\begin{tabular}{c@{\hspace{10ex}}c@{\hspace{10ex}}c}
\includegraphics[width=0.25\linewidth]{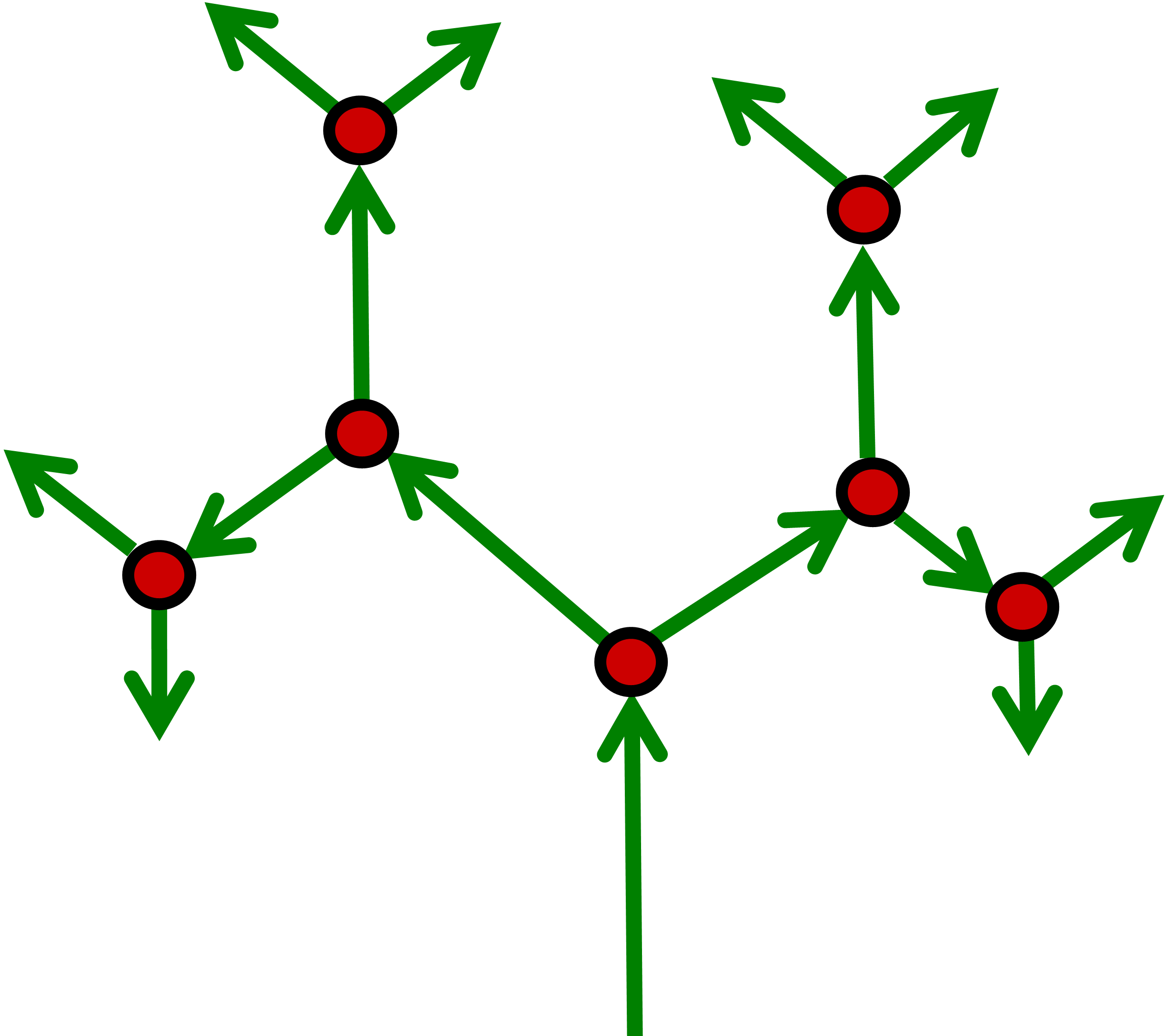} &
\includegraphics[width=0.25\linewidth]{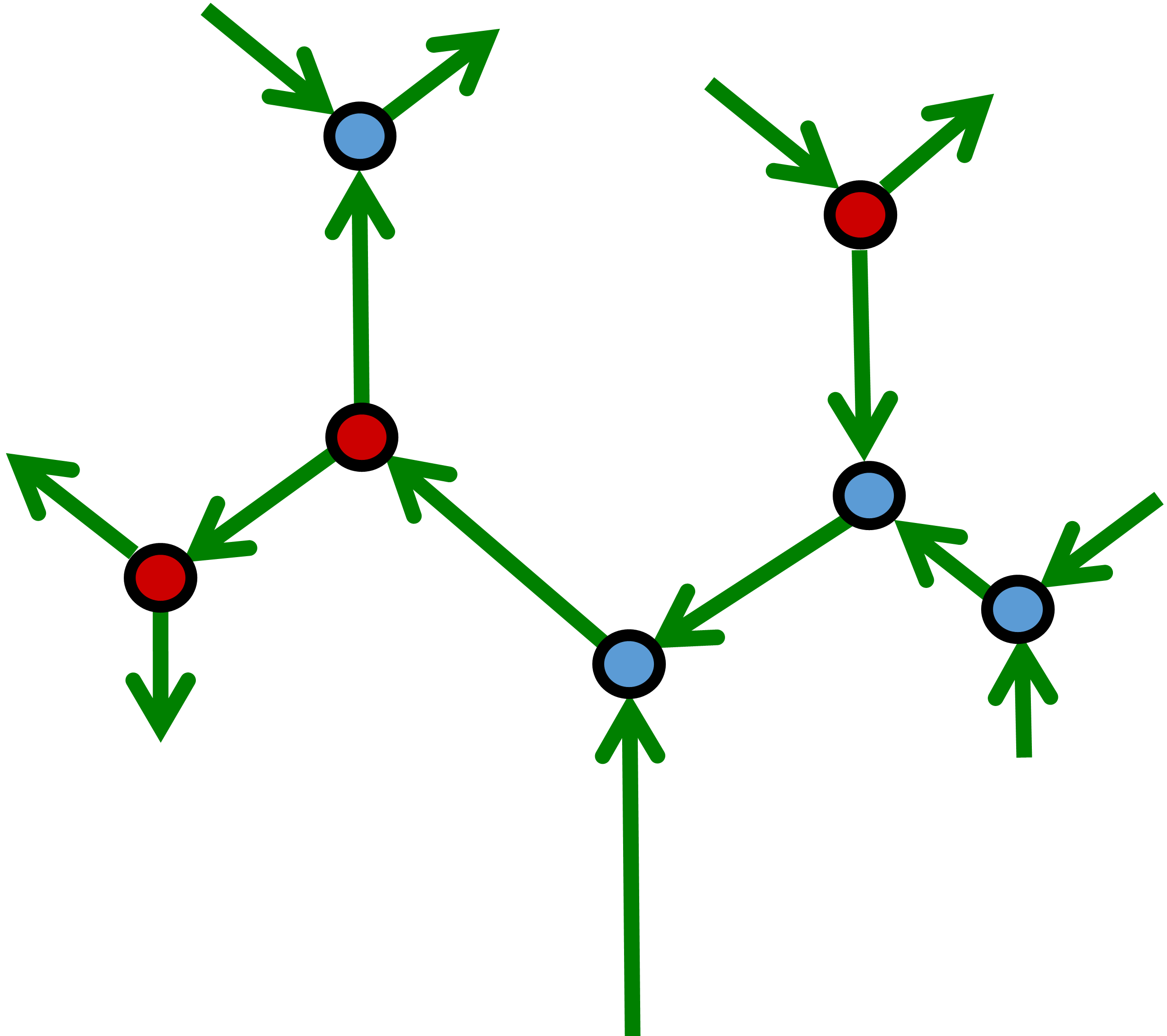} &
\includegraphics[width=0.25\linewidth]{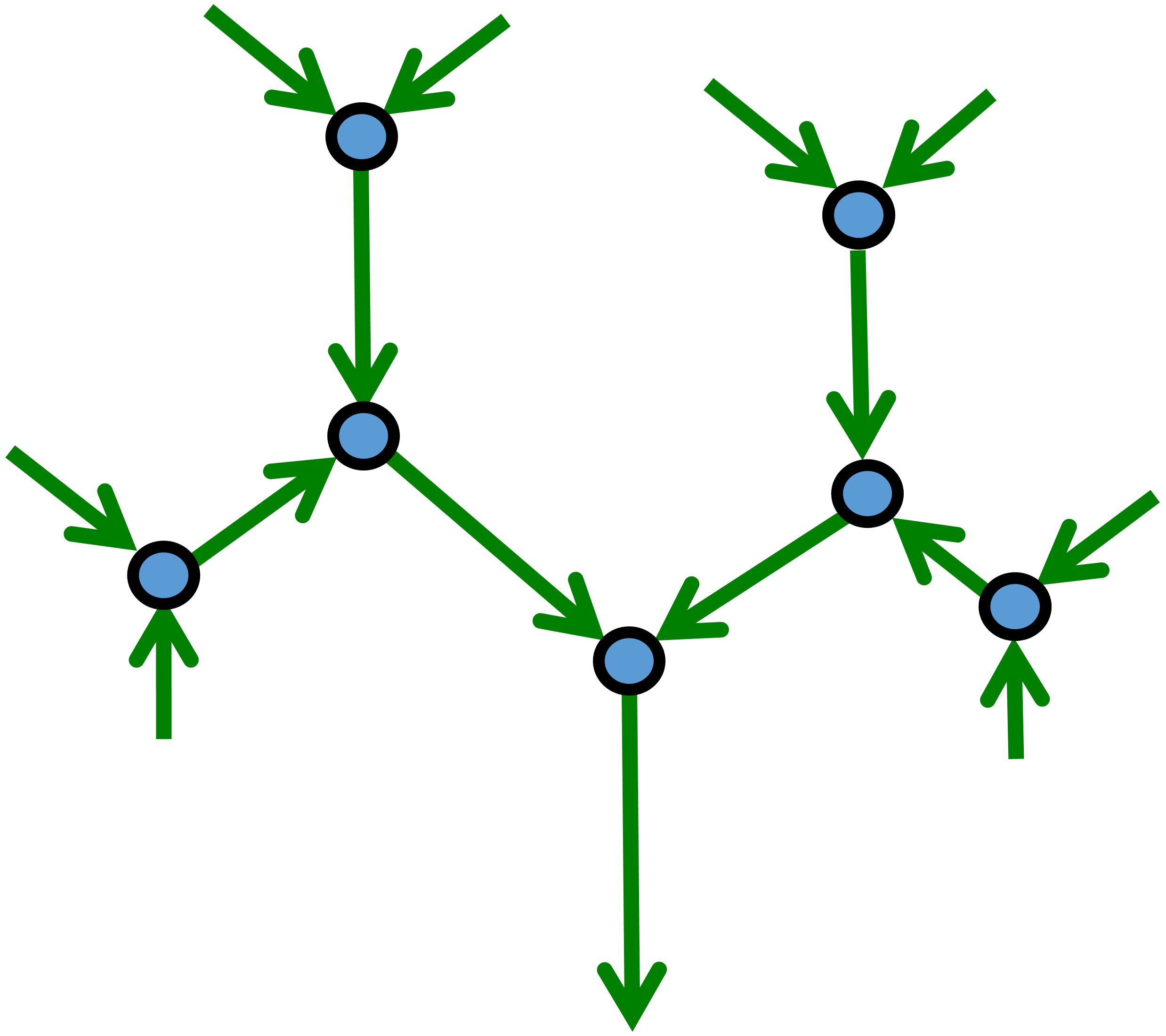} \\
(a) divergent vessels ({\bf arteries}) & (b) inconsistent divergence & (c) convergent vessels ({\bf veins}) \\[1ex]
\end{tabular}
\caption{{\bf [Vessel-tree divergence]} Vessels are the blood flow  {\em pathlines} 
and could be assigned orientations \eqref{eq:toward oriented}. To estimate orientations, we
penalilze negative (or positive) ``vessel divergence'', which we define as the divergence of oriented 
\underline{unit} tangents of vessels/pathlines. Such {\em unit tangent flow} divergence 
is positive (red) or negative (blue) at bifurcations, see (a-c).
Note that standard curvature  \cite{olsson2012curvature,thin:iccv15} and oriented 
curvature models \eqref{eq:oricurvature} either can not distinguish (b) from (a) and (c) or 
may even prefer (b) depending on specific combinations of bifurcation angles. For example, compare vessel direction
disambiguation based on curvature and divergence prior 
in Fig.\ref{fig:inconsistentorientation} (a) and (b). \label{fig:tree_divergence}}
\end{figure*}

This work addresses an important limitatation of vessel tree reconstruction methods
due to sign ambiguity in vessel orientation produced by local vesselness filters, e.g. Frangi. 
This orientation is described by the smallest eigen vector of the local intensity Hessian, but its sign
is ambiguous. Thus, the actial flow directions are not known, 
eventhough they are an important reconstruction cue particualrly at bifurcations.
This binary direction ambiguity can be resolved only by looking at the global configuration 
of vessel orientations (tangents) allowing to determine a consistent flow pattern.

\begin{figure}[t]
\centering
\small
\begin{tabular}{cc}
\includegraphics[width=0.45\linewidth,page=1]{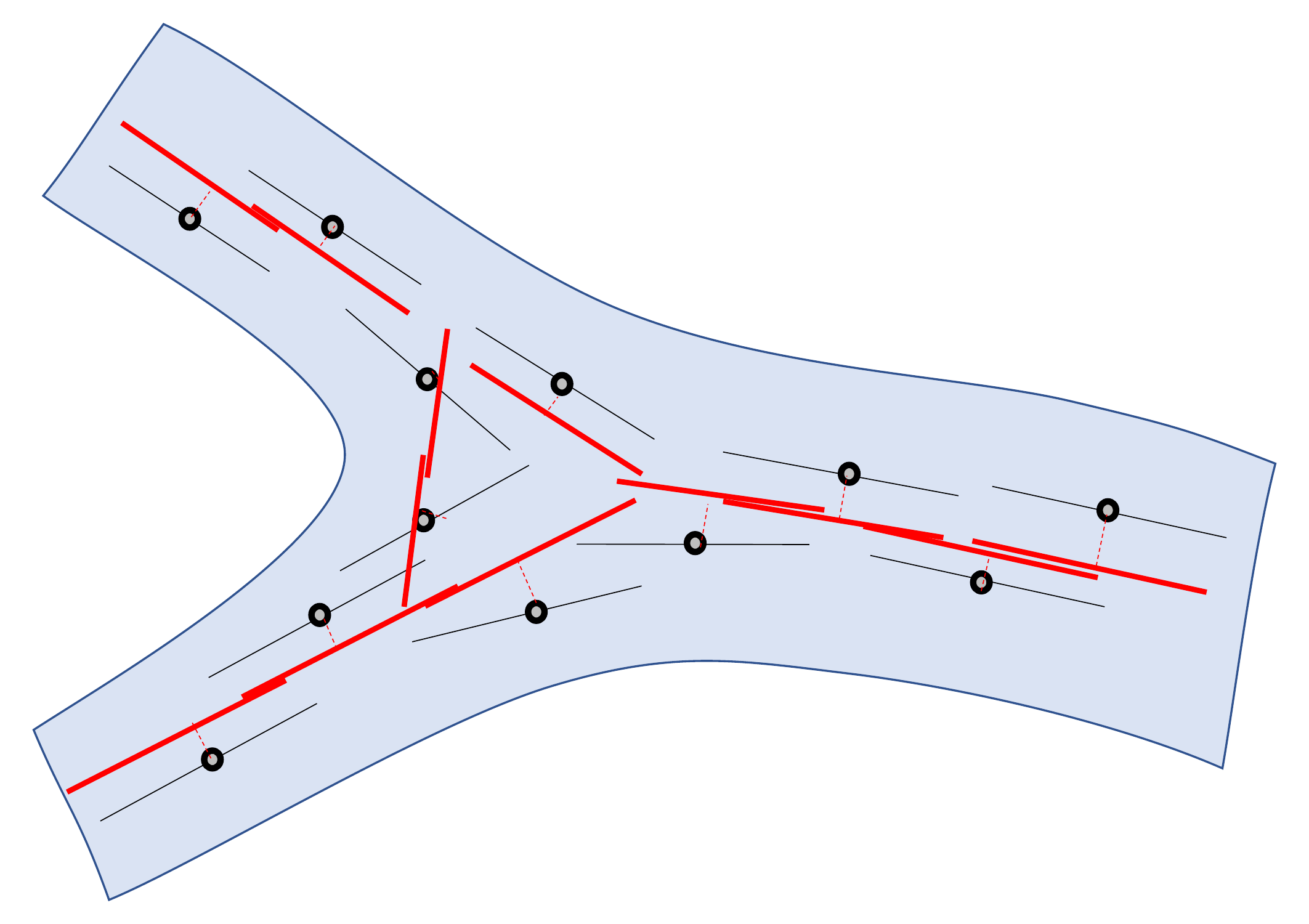} &
\includegraphics[width=0.45\linewidth,page=2]{fig/bifurcation.pdf} \\[-1ex]
(a) & (b)\\
\includegraphics[width=0.45\linewidth,height=0.2\linewidth]{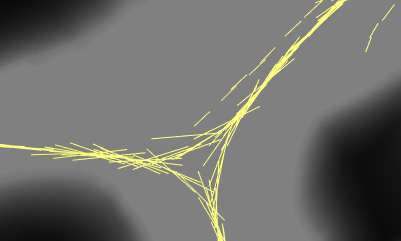} &
\includegraphics[width=0.45\linewidth,height=0.2\linewidth]{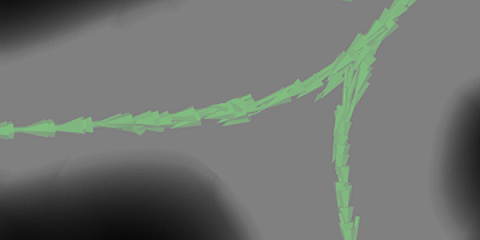}\\[-1ex]
(c) & (d)
\end{tabular}
\caption{Triangle artifacts at bifurcation. Optimization of energy \eqref{eq:unoriented curv energy} ignoring tangent orintations often leads to a strong local minima as in (a) and (c). The line segments are the estimated tangents of the centerline. New curvature term \eqref{eq:oricurvature} takes into account tangent orientations resolving the artifacts, see (b) and (d).}
\label{fig:triangleArtifact}
\end{figure}

We propose a divergence prior for disambiguating the global flow pattern over the vessel tree, 
see Figure \ref{fig:tree_divergence}. This prior can be imposed as a regularization
constraint for a vector field of oriented unit tangents for vessel pathlines. 
We penalize negative (or positive) divergence for such unit {\em tangent flow}
to enforce a consistent flow pattern\footnote{This divergence constraint is specific to unit tangent flow. 
Note that divergence for consistent blood flow velocities is zero even 
at bifurcations assuming {\em incompressible} blood. }\@.
The summary of our contributions:
\begin{itemize}
\item Prior knowledge about divergence is generally useful for vector field inference.
We propose a way to evaluate divergence for sparsely sampled vector fields via pairwise potentials. 
This makes divergence constraints amenable to a wide range of optimization methods for disrcrete of continuous hidden variables.
\item As an important application, we show that known divergence can disambiguate vessel directions produced by 
standard vessel filters, \eg Frangi \cite{frangi1998multiscale}. This requires estimation of binary ``sign'' variables. 
The constraint penalizing positive (or negative) divergence is non-submodular, but it is well optimized by TRWS \cite{trws}.
\item To estimate vessel tree centerline, divergence constraint can be combined with robust {\em oriented} curvature regularization for pathline tangents.
Additional options include outlier/detection variables \cite{thin:iccv15} and/or tree structure completion techniques, \eg using MST. 
\item We provide extensive quantitative validation on synthetic vessel data, as well as qualitative results on real high-resolution volumes.
\end{itemize}
The paper is organized as follows. Section \ref{sec:bifurcations} introduces oriented vessel pathline tangents
and discusses their curvature-based regularization. It is clear that orientation of the flow at the bifurcations
is important, \eg see Fig.\ref{fig:triangleArtifact}. Section \ref{sec:divergence} introduces our divergence prior and methods for 
enforcing it in the context of vessel tree centerline estimation. The last sections presents our experimental results.

\section{Bifurcations and curvature}
\label{sec:bifurcations}

\subsection{Oriented curvature constraint}
\label{sec:oriented_curvature}

\begin{figure}
\centering
\begin{tabular}{cm{0.85\linewidth}}
(a) & \includegraphics[width=\linewidth,page=1]{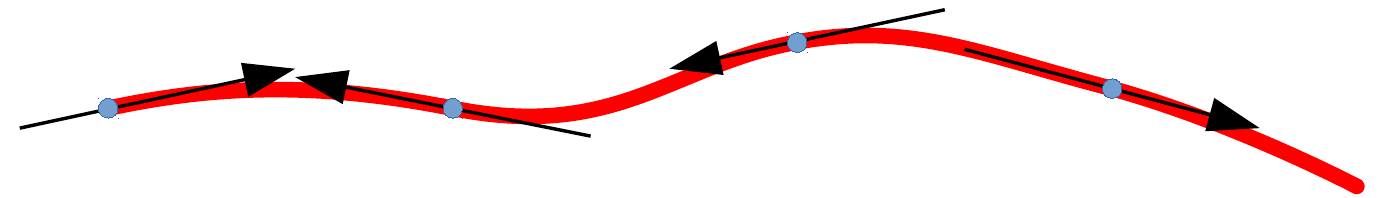} \\
(b) & \includegraphics[width=\linewidth,page=2]{fig/oriented_curvature2.pdf} \\
(c) & \includegraphics[width=\linewidth,page=3]{fig/oriented_curvature2.pdf} \\
\end{tabular}
\caption{The difference between unoriented (a) and oriented (b,~c) tangents.
Unoriented models ignore directions of tangents (a). Red color illustrates curves 
that comply with tangents in unoriented (a) and oriented (b,~c) cases.
Curvature approximations \mbox{(\ref{eq:abs curv}--\ref{eq:sqr curv})} are 
not able to distinguish (a), (b) and (c). Our oriented curvature \eqref{eq:oricurvature}
prefers configuration (c) over (a) and (b).}
\label{fig:oriented curvature toy}
\end{figure}

\begin{figure}
\centering
\begin{tabular}{m{0.3\linewidth}m{0.3\linewidth}m{0.3\linewidth}}
\includegraphics[width=\linewidth,page=1]{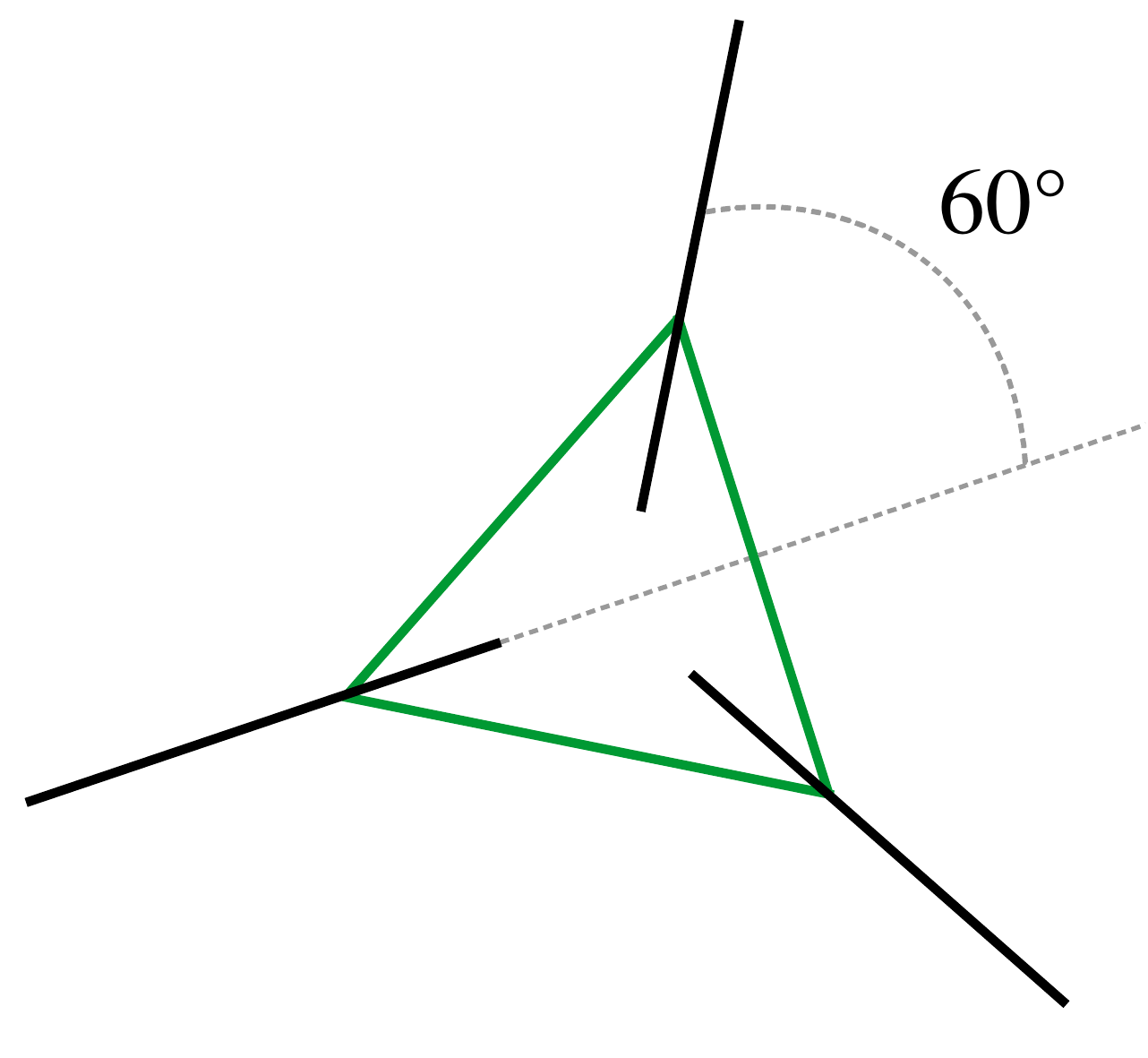} &
\includegraphics[width=\linewidth,page=2]{fig/triangles.pdf} &
\includegraphics[width=\linewidth,page=3]{fig/triangles.pdf} \\[-3ex]
\centering (a) &\centering  (b) &\centering  (c) \\[1ex]
\end{tabular}
\caption{Illustrative examples of three interacting tangents with unoriented curvature (a) as in energy \eqref{eq:unoriented curv energy} and two alternative oriented configurations (b) and (c) with oriented curvature as in energy \eqref{eq:oriented curv energy}. The green line denotes pairwise interaction with low curvature value. Note, that unoriented curvature (\ref{eq:abs curv}--\ref{eq:sqr curv}) always chooses the smallest angle for calculation. The red line shows ``inactive'' pairwise interaction, \ie interaction where curvature in \eqref{eq:oricurvature} reaches the high saturation threshold. }
\label{fig:triangles toy}
\end{figure}

Previous works \cite{olsson2012curvature,olsson2013partial,thin:iccv15}
ignored orientations of tangent vectors $\{l_p\}_{p \in \Omega}$. 
Equations \eqref{eq:abs curv}--\eqref{eq:unoriented curv energy} 
do not depend on orientations of $l$. 
In practice, the orientations of vectors $l_p$ are arbitrarily defined.
Ingnoring the orientations in energy \eqref{eq:unoriented curv energy}
results in significant ``triangle'' artifacts around bifurcation, see Fig.~\ref{fig:triangleArtifact}(a,c). 
Consider an illustrative example in Fig.~\ref{fig:triangles toy}(a). 
Each of three tangents interacts with the other two.  
The prior knowledge about blood flow pattern dictates that among those three tangents 
there should be one incoming and one outcoming. Introduction of orientations allows
us to distinguish the incoming/outcoming tangents and
subsequently inactivate one of the interactions, see Fig.~\ref{fig:triangles toy}(b), 
resulting in disappearance of these artifacts.

\begin{figure}[h!]
\begin{tabular}{m{0.45\linewidth}m{0.45\linewidth}}
\caption{Robustness of curvature \eqref{eq:oricurvature}.
The pairs of tangent vectors that has angle greater than 
$\operatorname{acos}\tau$ are not considered belonging 
to the same vessel. A constant penalty is assigned to such pairs.
This ``turns off'' smoothness enforcement at bifurcations. 
\label{fig:robust} }&
\includegraphics[width=\linewidth]{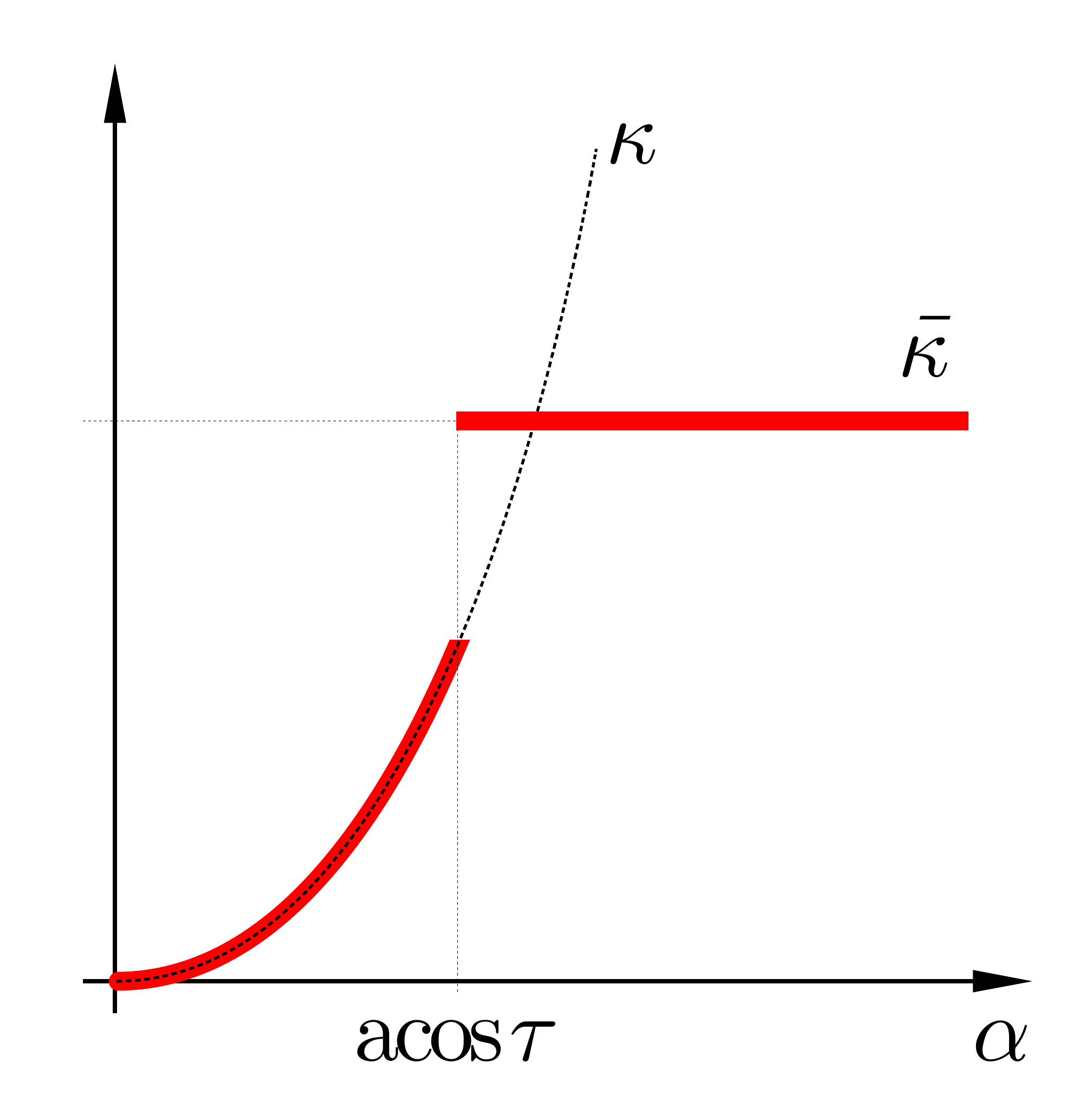}
\end{tabular}
\end{figure}

In order to introduce oriented curvature we introduce a new vector field $\bar l_p$, which we call \emph{oriented}. Then, we introduce  energy $E_o(\bar l)$ by replacing curvature term in energy \eqref{eq:unoriented curv energy} with a new oriented curvature as follows
\begin{equation}\label{eq:oriented curv energy}
E_o(\bar l) = \sum_p{\| \tilde p - \bar l_p \|^2} + \gamma \sum_{(p, q)\in N}\bar\kappa_{pq}(\bar l_p, \bar l_q)
\end{equation}
where 
\begin{equation} \label{eq:oricurvature} 
\bar\kappa_{pq}(\bar l_p, \bar l_q) \;\; := \;\; \begin{cases}
\kappa_{pq}(\bar l_p, \bar l_q), & \langle \bar l_p, \bar l_q\rangle \ge \tau, \\
1, & \text{otherwise,}
\end{cases}
\end{equation}
and $ \langle \bar l_p, \bar l_q \rangle$ is the dot product of $\bar l_p$ and $\bar l_q$ 
and $\tau \ge 0$ is a positive threshold discussed in Fig.~\ref{fig:robust}.

The connection between oriented field $\bar l$ and $l$ is
\begin{equation} \label{eq:toward oriented}
\bar l_p = x_p \cdot l_p
\end{equation}
where binary variables $x_p \in \{ -1, 1 \}$ flip or preserve the arbitrarily defined orientations of $l_p$.  

\subsection{Curvature and orientation ambiguity}
\label{sec:Xestimation}

\begin{figure}[t]
\centering
\begin{tabular}{c}
\includegraphics[width=\linewidth]{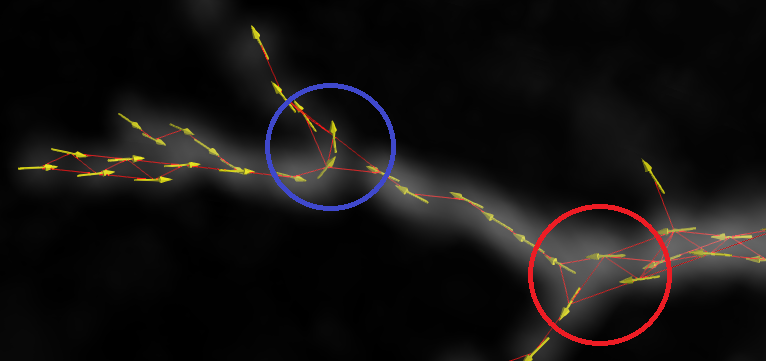} 
\\[-1ex] (a) oriented curvature only \eqref{eq:orient energy} \\
\includegraphics[width=\linewidth]{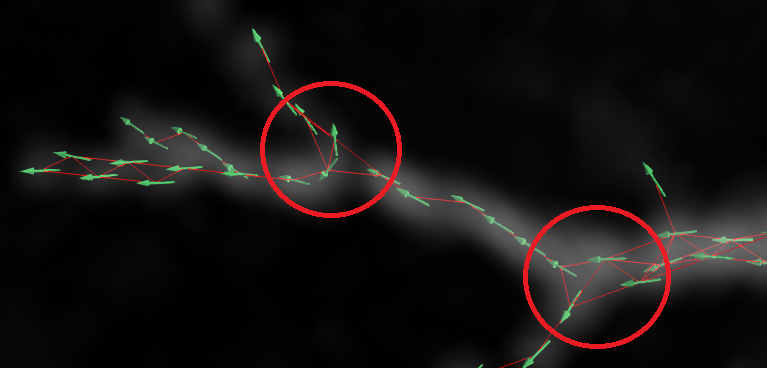} 
\\[-1ex] (b) with divergence prior \eqref{eq:sign energy} 
\end{tabular}
\caption{Disambiguating flow directions in Frangi output \cite{frangi1998multiscale}.
Both examples use fixed (unoriented) vessel tangents $\{l_p\}$ produced by the filter and
compute (oriented) vectors $\bar{l}_p = x_p l_p$ \eqref{eq:toward oriented} by optimizing binary sign variables $\{x_p\}$
using energies \eqref{eq:orient energy} in (a) and \eqref{eq:sign energy} in (b). The circles indicate
divergent (red) or convergent (blue) bifurcations similarly to the diagrams in Fig.\ref{fig:tree_divergence}.
The extra divergence constraint in \eqref{eq:joint energy}  enforces consistent flow pattern (b).}
\label{fig:inconsistentorientation}
\end{figure}

Introduction of orientated curvature resolves triangle artifacts, 
see Fig.~\ref{fig:triangleArtifact}(b,d). However, the orientations are not known in advance. 
For example, Frangi filter \cite{frangi1998multiscale} defines
a tangent as a unit eigen vector of a special matrix. 
The unit eigen vectors are defined up to orientation, which is chosen
arbitrarily. One may propose to treat energy \eqref{eq:oriented curv energy} 
as a function of tangent orientations $x$ via relationship \eqref{eq:toward oriented} as follows
\begin{equation}\label{eq:orient energy}
E_o(x) \;\; := \;\; E_o(\{x_p \cdot l_p\}) \Big|_{l_p=const}
\end{equation}
However, energy \eqref{eq:orient energy} is under-constrained because 
it allows multiple equally good solutions,  see Fig.~\ref{fig:triangles toy}(b) and (c). 
The example in (b) shows a divergent pattern while
(c) shows a convergent pattern suggesting artery/vein ambiguity. 
Unfortunately, energy \eqref{eq:orient energy} does not enforce  consistent flow pattern across 
the vessel tree resulting in a mix of divergent and convergent bifurcations as in Fig.~\ref{fig:tree_divergence}(b).
Real data experiments confirm this conclusion, see Fig.~\ref{fig:inconsistentorientation}(a).

Thus, oriented curvature model \eqref{eq:oriented curv energy} has a significant problem.
While it can resolve ``triangle artifacts'' at bifurcations, see Fig.\ref{fig:triangleArtifact},
it will break the wrong sides of the triangles at many bifurcations where it estimates the flow pattern incorrectly and then give the incorrect estimation of centerline, see Fig.\ref{fig:inconsistentorientationcenterline}(a). 
Below we introduce our divergence prior directly enforcing consistent flow pattern over the vessel tree.

\section{Divergence constraint}
\label{sec:divergence}

\subsection{Estimating divergence}
\label{sec:d_estimation}

\begin{figure}[h]
\centering
\begin{tabular}{c}
\includegraphics[width=2.1in]{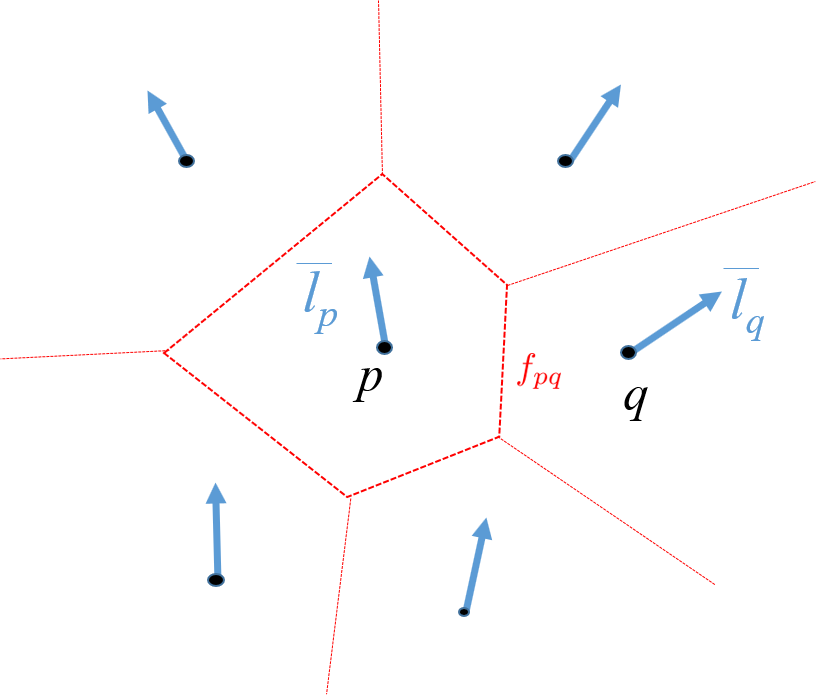}   \\ (a)  Voronoi cells for $p,q\in\Omega$ and facet $f_{pq}$  \\[1ex]
\includegraphics[width=2.1in]{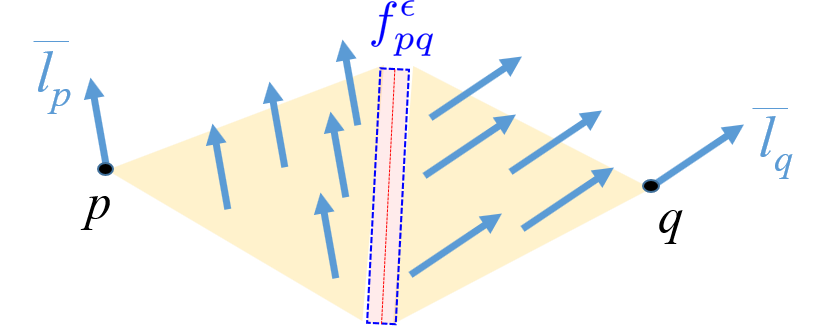}    \\  (b) $\epsilon$-thin box $f^{\epsilon}_{pq}$ around facet $f_{pq}$ \\[1ex]
\end{tabular}
\caption{Divergence of a sparse vector field $\{\bar{l}_p|p\in\Omega\}$. Assuming that the corresponding 
``extrapolated'' dense vector field is constant inside Voronoi cells (a), 
it is easy to estimate (non-zero) divergence $\nabla \bar{l}_{pq}$ \eqref{eq:flux} concentrated in a narrow region $f^{\epsilon}_{pq}$ 
around each facet (b) using the divergence theorem. } \label{fig:divergence}
\end{figure}

Figure \ref{fig:divergence} describes our (finite element) model for estimating divergence of a sparse vector field
$\{\bar{l}_p|p\in\Omega\}$ defined for a finite set of points $\Omega\subset{\cal R}^3$. We extrapolate the vector field 
over the whole domain ${\cal R}^3$ assuming constancy of the vectors on the interior of the Voronoi cells 
for $p\in\Omega$, see Fig.\ref{fig:divergence}(a).
Thus, vectors change only in the (narrow) region around the cell facets where all non-zero divergence is concentrated. 
To compute the integral of divergence in the area between two neighboring points $p,q\in\Omega$, see Fig.\ref{fig:divergence}(b),
we estimate flux of the extrapolated vector field over $\epsilon$-thin box $f^{\epsilon}_{pq}$ around facet $f_{pq}$
$$
\int_{f^{\epsilon}_{pq} } \langle \bar{l},n_s\rangle \; ds \;\;=\;\; 
\frac{\langle \bar{l}_q,pq \rangle - \langle \bar{l}_p,pq \rangle}{|pq|}\cdot |f_{pq}| \;\;+\;\; o(\epsilon)
$$
where $n_s$ is the outward unit normal of the box and $|f_{pq}|$ is the facet's area.
Then, divergence theorem implies the following formula for the integral of divergence of the vector field inside box $f^{\epsilon}_{pq}$
\begin{equation} \label{eq:flux}
\nabla \bar{l}_{pq}\;\;\ = \;\; 
\frac{\langle \bar{l}_q,pq \rangle - \langle \bar{l}_p,pq \rangle}{|pq|}\cdot |f_{pq}|
\end{equation}
where we ignore only infinitesimally negligible $o(\epsilon)$ term. 

\begin{figure}[t]
\centering
\begin{tabular}{c}
\includegraphics[width=\linewidth]{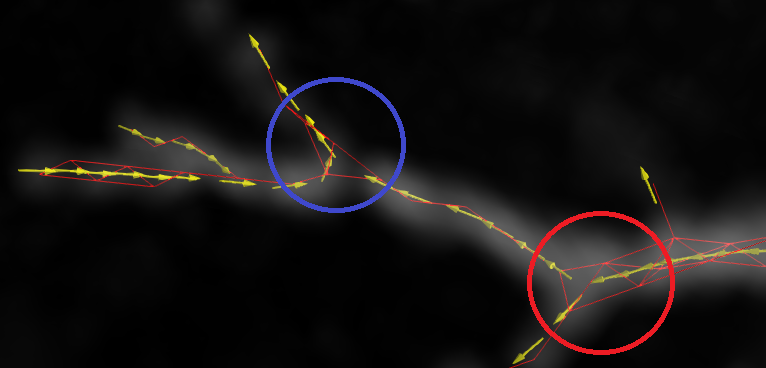}
\\[-1ex] (a) tangent vectors at convergence for energy \eqref{eq:oriented curv energy}\\
\includegraphics[width=\linewidth]{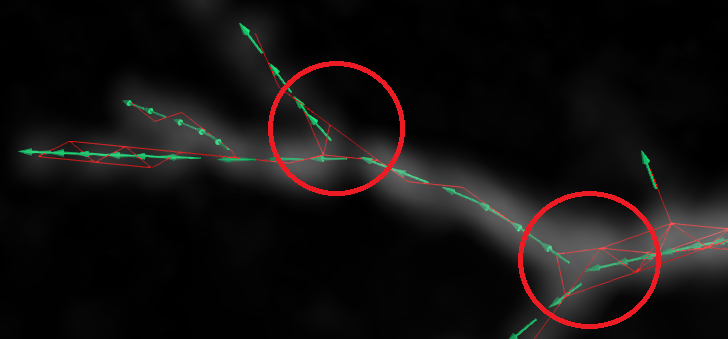}
\\[-1ex] (b) tangent vectors at convergence for energy \eqref{eq:joint energy} 
\end{tabular}
\caption{Centerline estimation for the data in Fig.\ref{fig:inconsistentorientation}. Instead of showing tangent
orientations estimated at the first iteration as in Fig.\ref{fig:inconsistentorientation}, 
we now show the final result at convergence for minimizing energy 
\eqref{eq:oriented curv energy} in (a) and energy \eqref{eq:joint energy}  in (b).
Blue circle shows bifurcation reconstruction artifacts due to wrong estimation of vessel orientations
in Fig.\ref{fig:inconsistentorientation}(a).}
\label{fig:inconsistentorientationcenterline}
\end{figure}

\subsection{Oriented centerline estimation}
\label{sec:joint_energy}

Constraints for divergence $\nabla \bar{l}_{pq}$ in the regions between neighbors $p,q\in{\cal D}$ 
in Delaugney triangulation of $\Omega$ can be combined with $E_o(\bar{l})$ in \eqref{eq:oriented curv energy} 
to obtain the following joint energy for estimating oriented centerline tangents $\bar{l}_p$
\begin{equation} \label{eq:joint energy}
E(\bar{l}) \;\;=\;\; E_o(\bar{l}) \;\;\;+\;\;\; \lambda \;\sum_{(p, q)\in {\cal D}}  (\nabla \bar{l}_{pq})^-
\end{equation}
where the negative part operator $(\cdot)^-$ encourages divergent flow pattern as in Fig.\ref{fig:tree_divergence}(a).
Alternatively, one can use $(\cdot)^+$ to encourage a convergent flow pattern as in Fig.\ref{fig:tree_divergence}(c).
This joint energy for oriented centerline estimation $E(\bar{l})$ combines Frangi measurements, 
centerline curvature regularity, and consistency of the flow pattern, see Fig.\ref{fig:inconsistentorientation}(b).
Note that specific value of facet size in \eqref{eq:flux} had a negligible effect in our centerline estimation tests
as it only changes a relative weight of the divergence penalty at any given location. For simplicity, one may use
$|f_{pq}|\approx const$ for all $p,q\in{\cal D}$.
\begin{figure}
\centering
\includegraphics[width=0.8\linewidth]{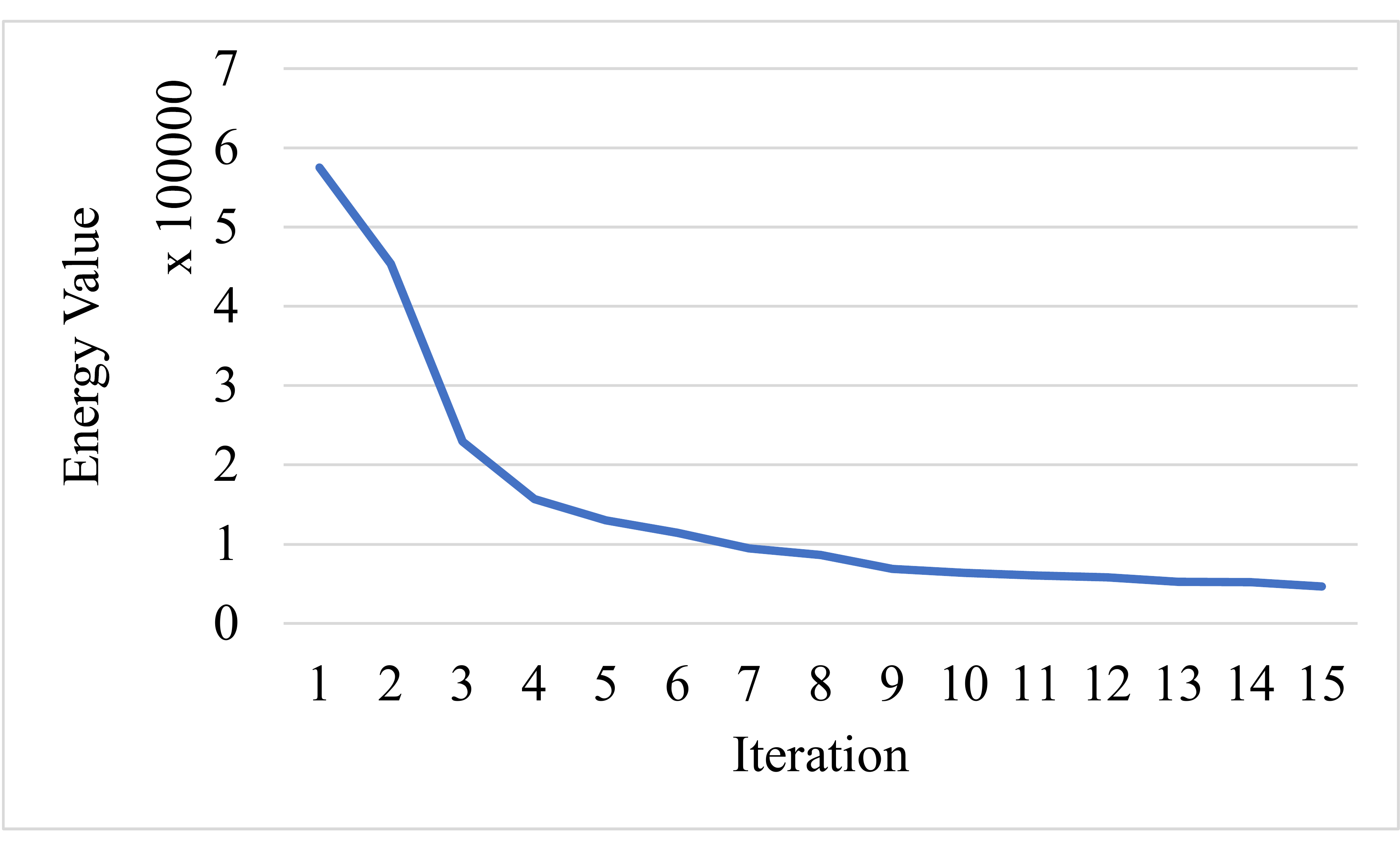}
\caption{Representative example of decrease in energy  \eqref{eq:joint energy}  
for block-coordinate descent iterating optimization of \eqref{eq:sign energy}  and \eqref{eq:un energy}.
For initialization, we use raw undirected tangents $\{l_p\}$ generated by Frangi filter \cite{frangi1998multiscale}.
Then, we iteratively reestimate binary sign variables $\{x_p\}$ and unoriented tangents $\{l_p\}$.}
\label{fig:convergencecurve}
\end{figure}

Optimization of oriented centerline energy $E(\bar{l})$ in \eqref{eq:joint energy} over oriented tangents $\{\bar{l}_p\}$ 
can be done via block-coordinate descent. As follows from definition \eqref{eq:toward oriented}
$$ E(\bar{l}) \;\;\equiv\;\; E(\{x_p\cdot l_p\}).$$
We iterate TRWS \cite{trws} for optimizing non-submodular energy for binary ``sign'' disambiguation variables $\{x_p\}$
\begin{equation} \label{eq:sign energy}
E(x) \;\; :=\;\; E(\{x_p\cdot l_p\})\Big|_{l_p=const}
\end{equation}
and {\em trust region} \cite{wright1985inexact,thin:iccv15} for optimizing robust energy for aligning tangents into 1D centerline
\begin{equation} \label{eq:un energy}
E(l) \;\; :=\;\; E(\{x_p\cdot l_p\})\Big|_{x_p=const} . 
\end{equation}
Figure \ref{fig:convergencecurve}
shows a representative example illustrating convergence of energy  \eqref{eq:joint energy} in a few iterations.

Note that the divergence constraint in joint energy  \eqref{eq:joint energy} 
resolves the problem of under-constrained objective \eqref{eq:oriented curv energy} 
discussed at the end of Section \ref{sec:bifurcations}. Since the flow pattern consistency is enforced,
optimization of \eqref{eq:joint energy} should lead to a consistent resolution of triangle artifacts at
bifurcations. see Fig.\ref{fig:inconsistentorientationcenterline}(b). Our experimental results support this claim. 

\section{Evaluation}
\label{sec:evaluation}

\subsection{Synthetic vessel volume}
\label{sec:synthetic}

\begin{figure}[t]
\centering
\includegraphics[width=0.8\linewidth]{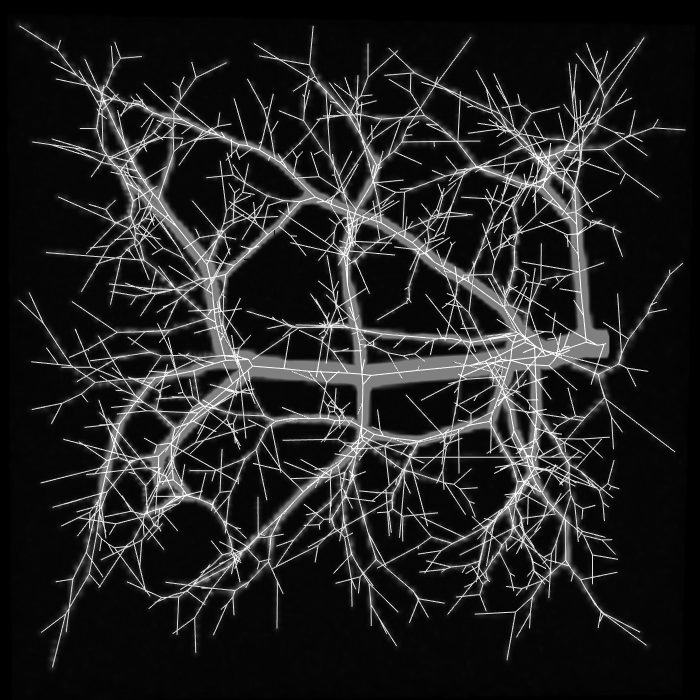}
\caption{An example of one volume synthetic data. The white lines inside vessels denote ground truth of centerline.}
\label{fig:wholeVolumeSynData}
\end{figure}

\begin{figure*}[t]
\centering
\includegraphics[width=0.33\linewidth]{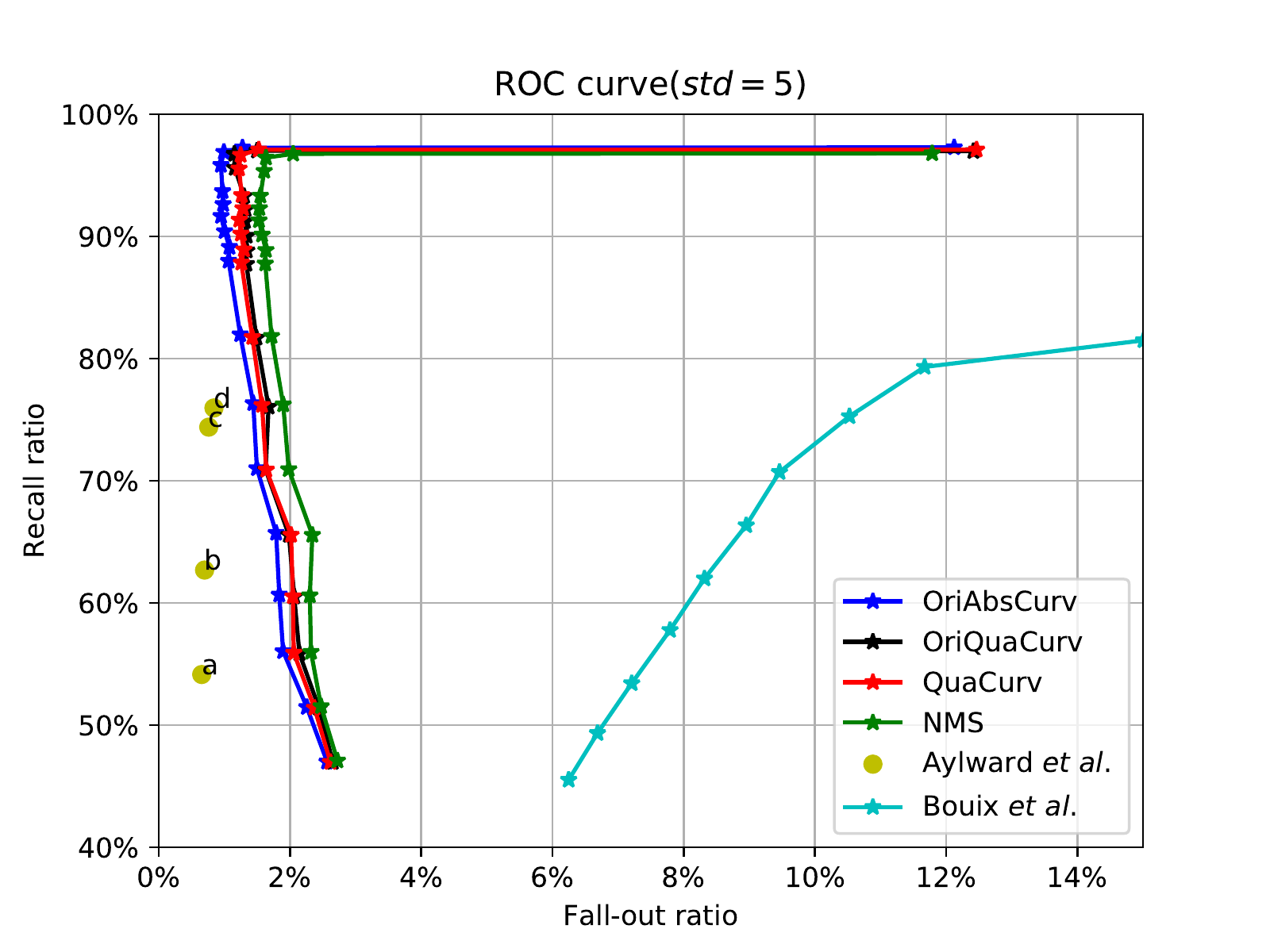}
\includegraphics[width=0.33\linewidth]{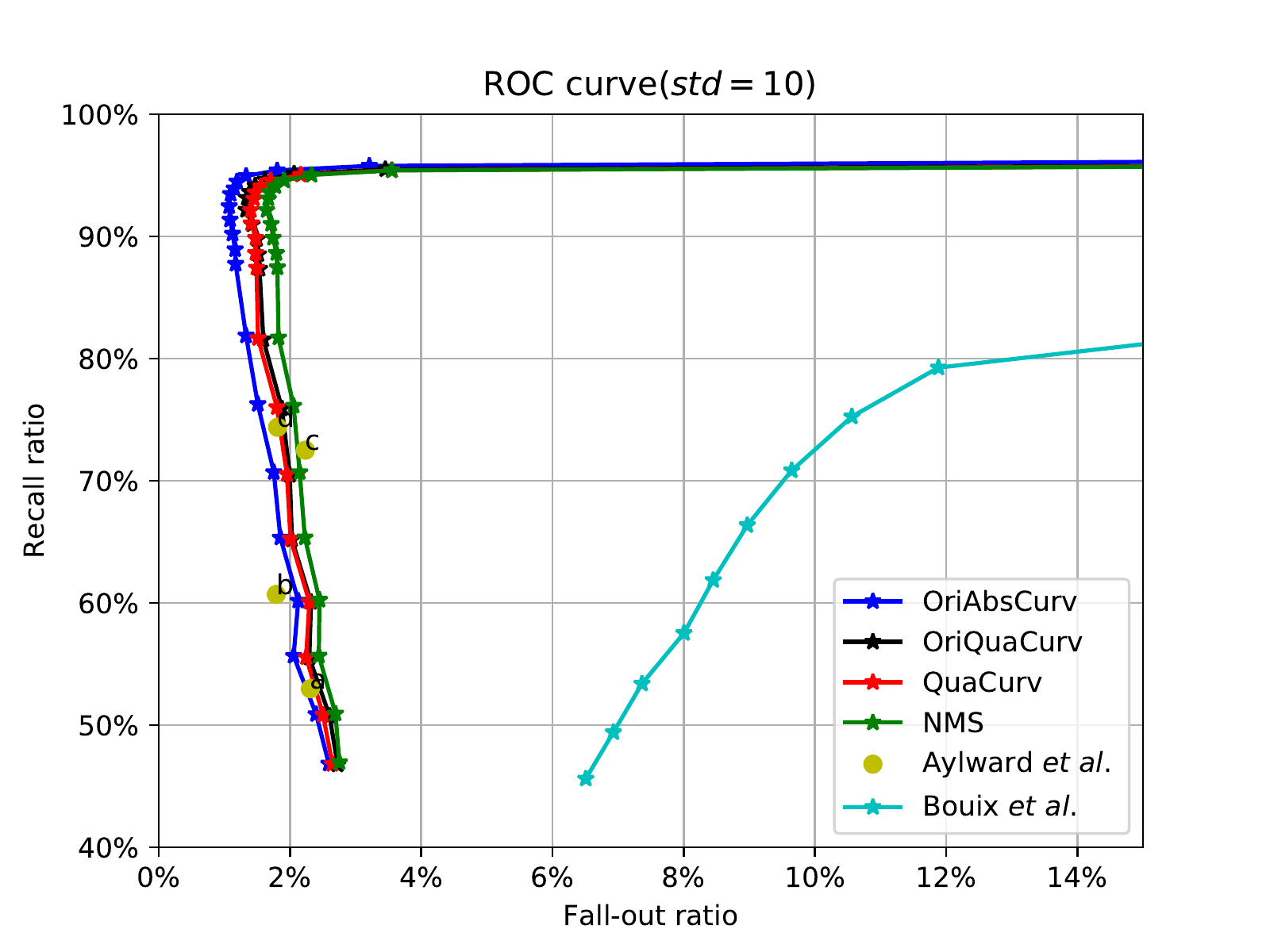}
\includegraphics[width=0.33\linewidth]{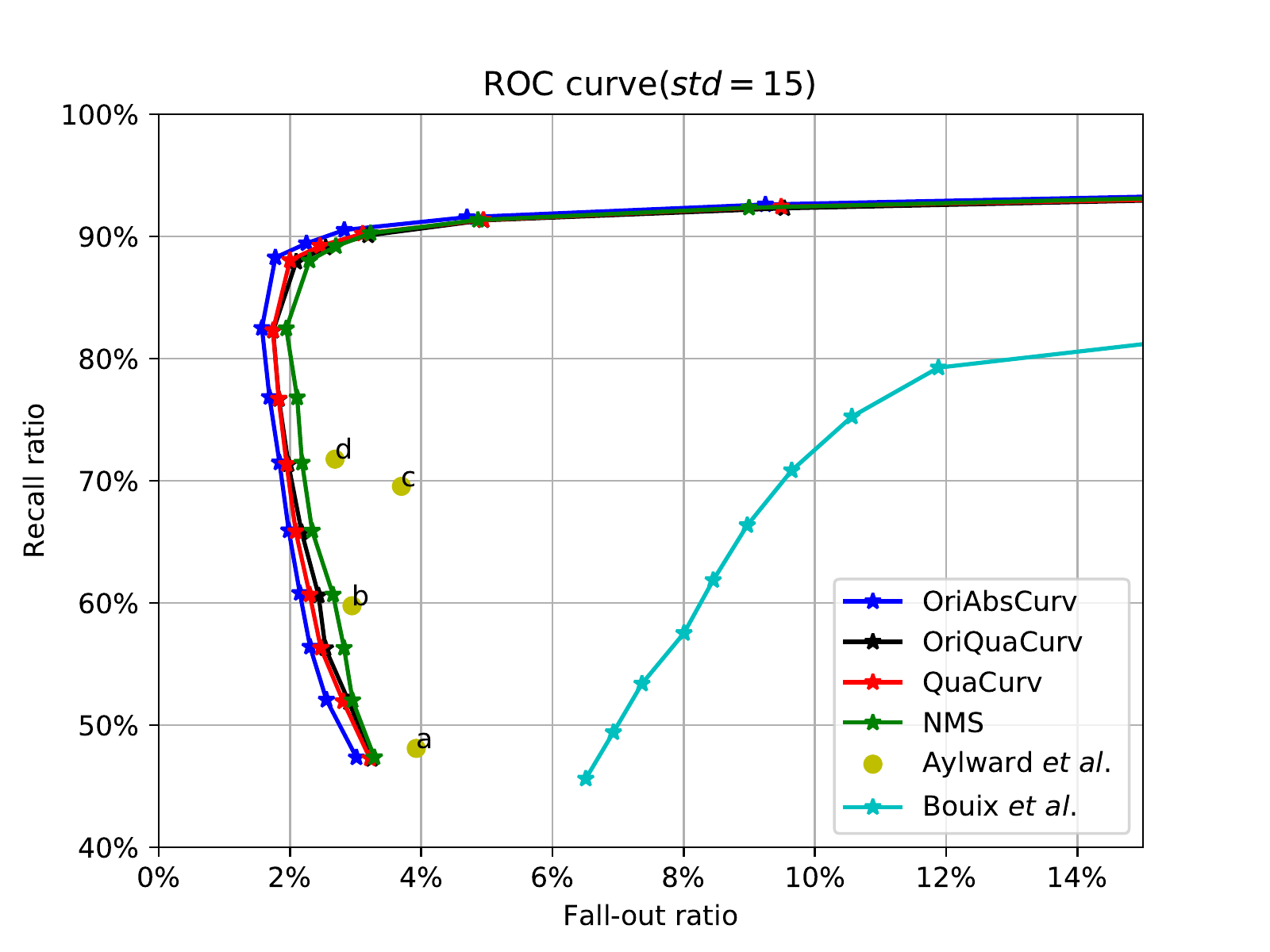}
\caption{Comparison of our method (OriAbsCurv and OriQuaCurv) with the unoriented quadratic curvature (QuaCurv) \cite{thin:iccv15}, non-maximum suppression (NMS), SegmentTubes (Aylward $et\ al.$ \cite{aylward2002initialization}) and medial axis extraction (Bouix $et\ al.$ \cite{bouix2005flux}) at three different noise levels. The four letters on yellow circles denote different seed point lists. $a$: using root and all leaf points; $b$: using 50\% of the mixture of all bifurcation and leaf points and root; $c$: using middle points of all branch segments; $d$: using all bifurcation and leaf points and root.}
\label{fig:rocResults}
\end{figure*}

\begin{figure*}[t]
\centering
\small
\includegraphics[width=0.33\linewidth]{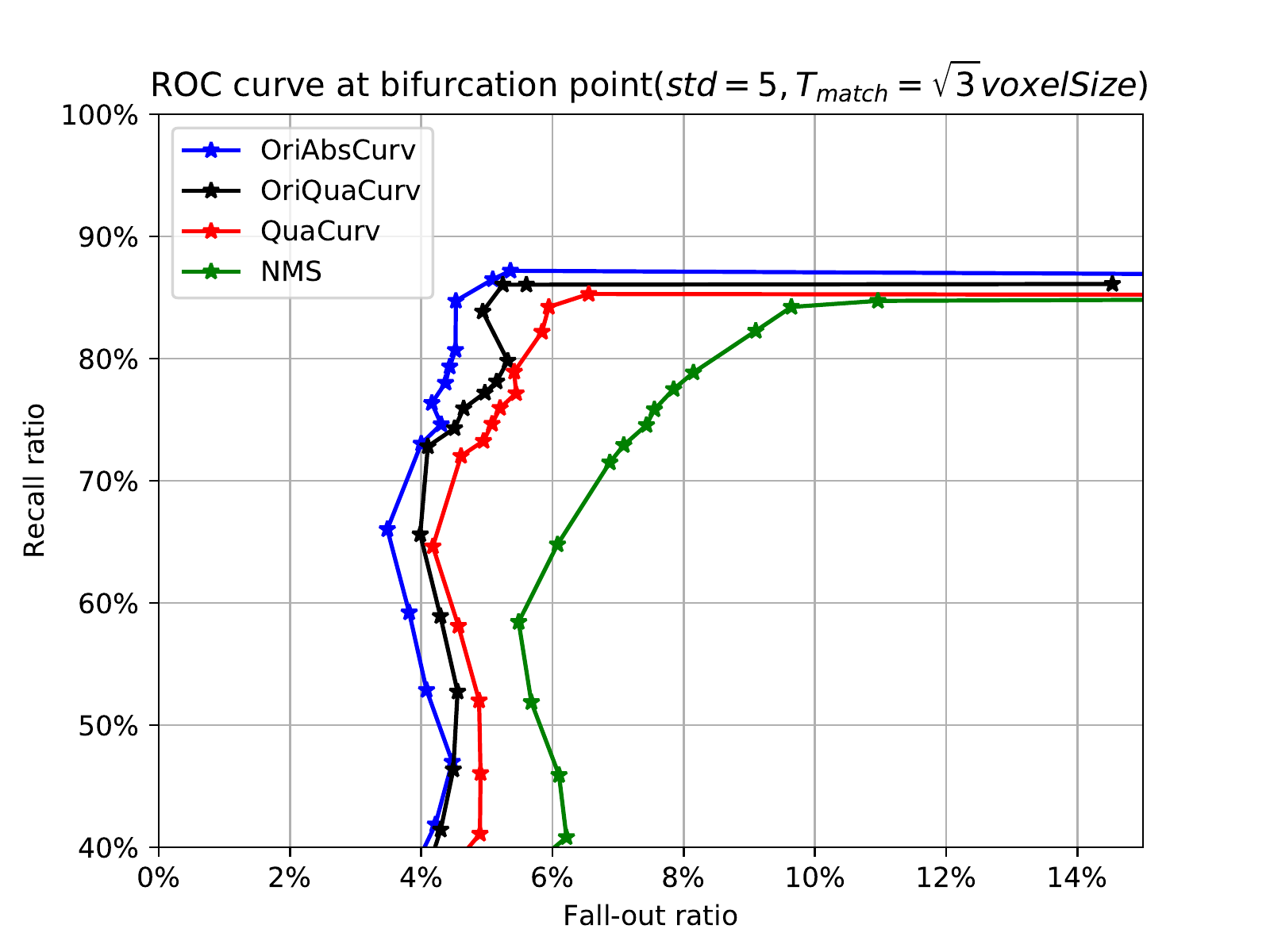}
\includegraphics[width=0.33\linewidth]{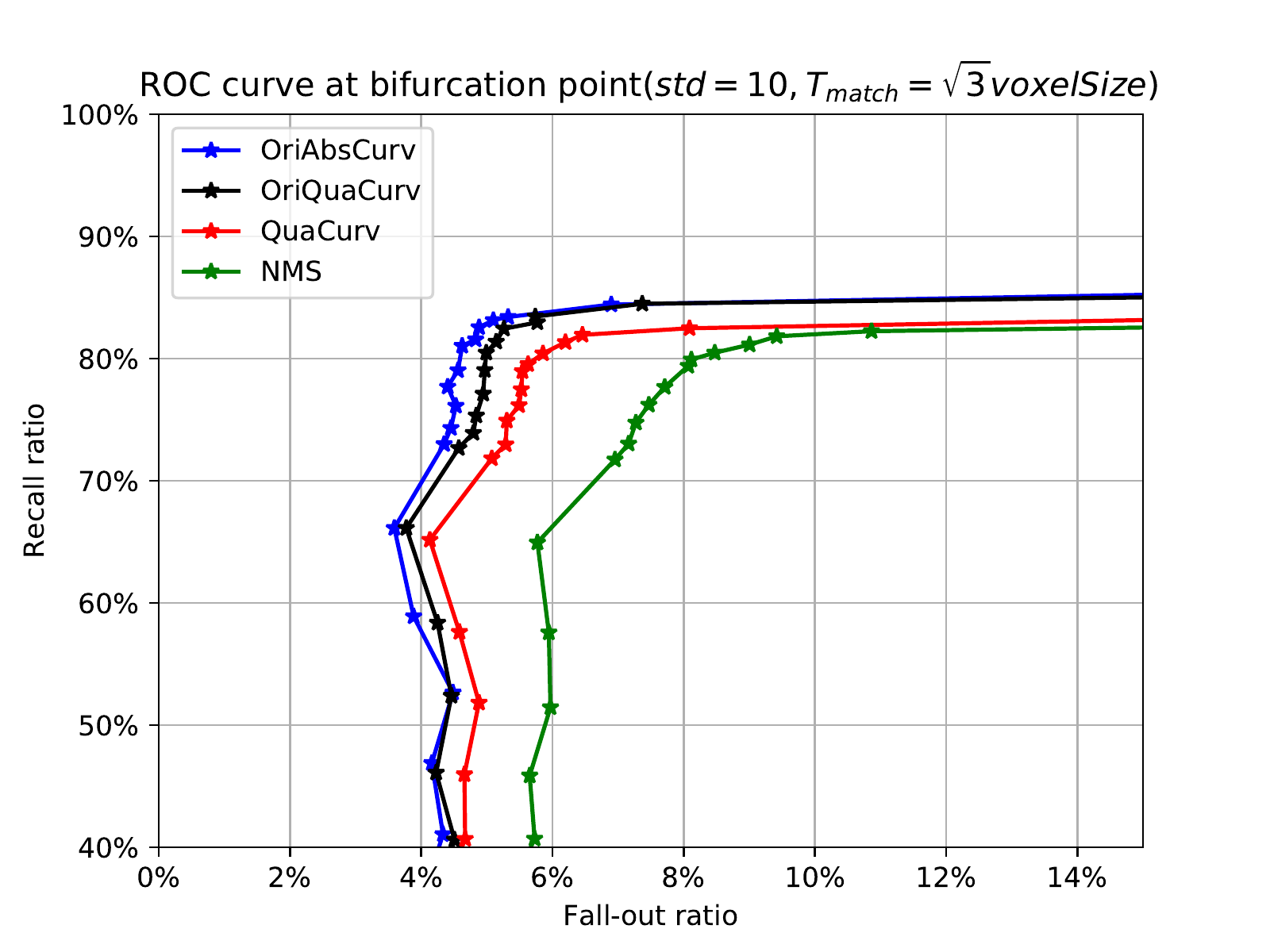}
\includegraphics[width=0.33\linewidth]{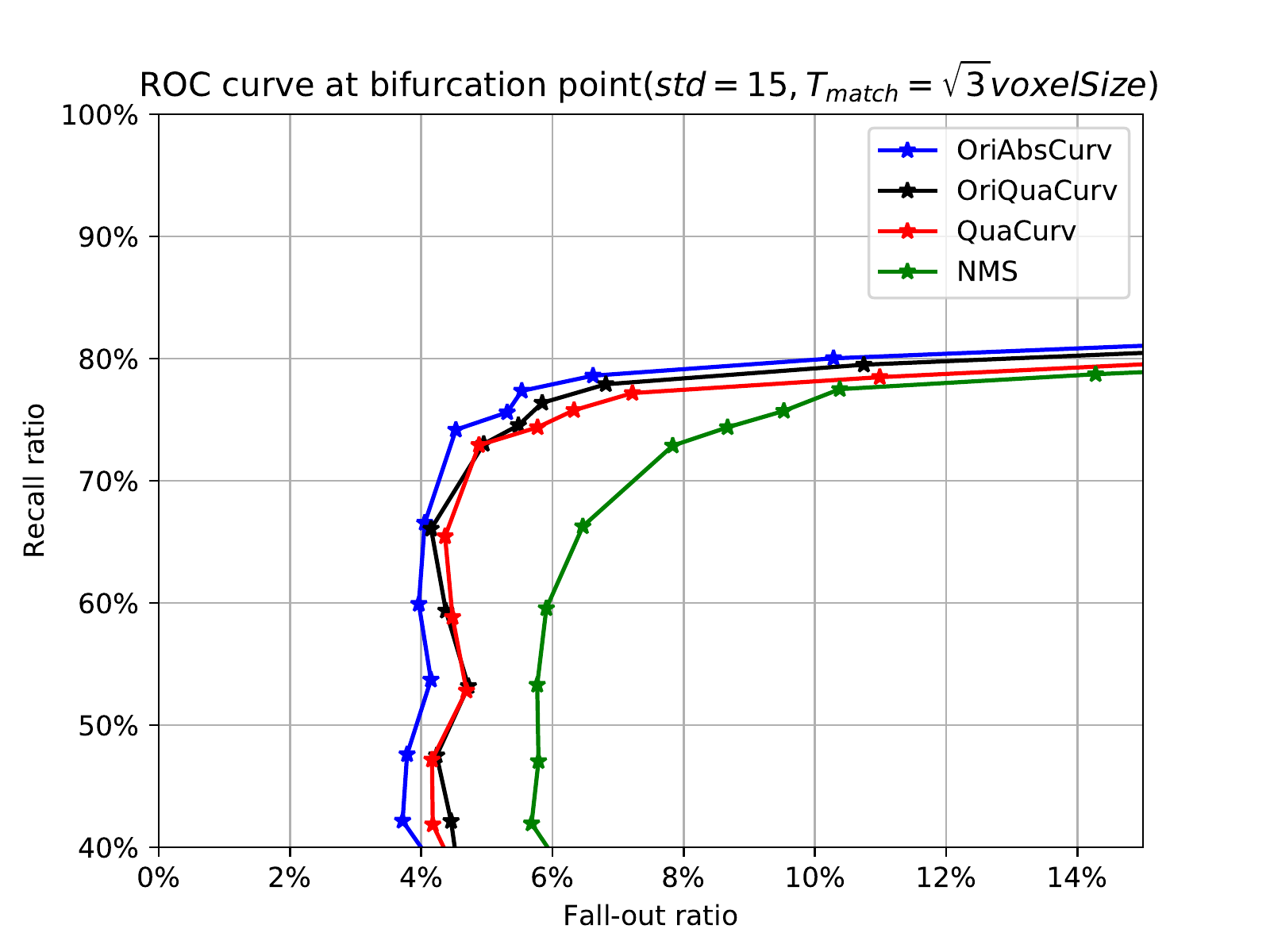}
\caption{Comparison only at bifurcation point. }
\label{fig:BifurRocResults}
\end{figure*}

\begin{figure*}[t]
\centering
\small
\includegraphics[width=0.33\linewidth]{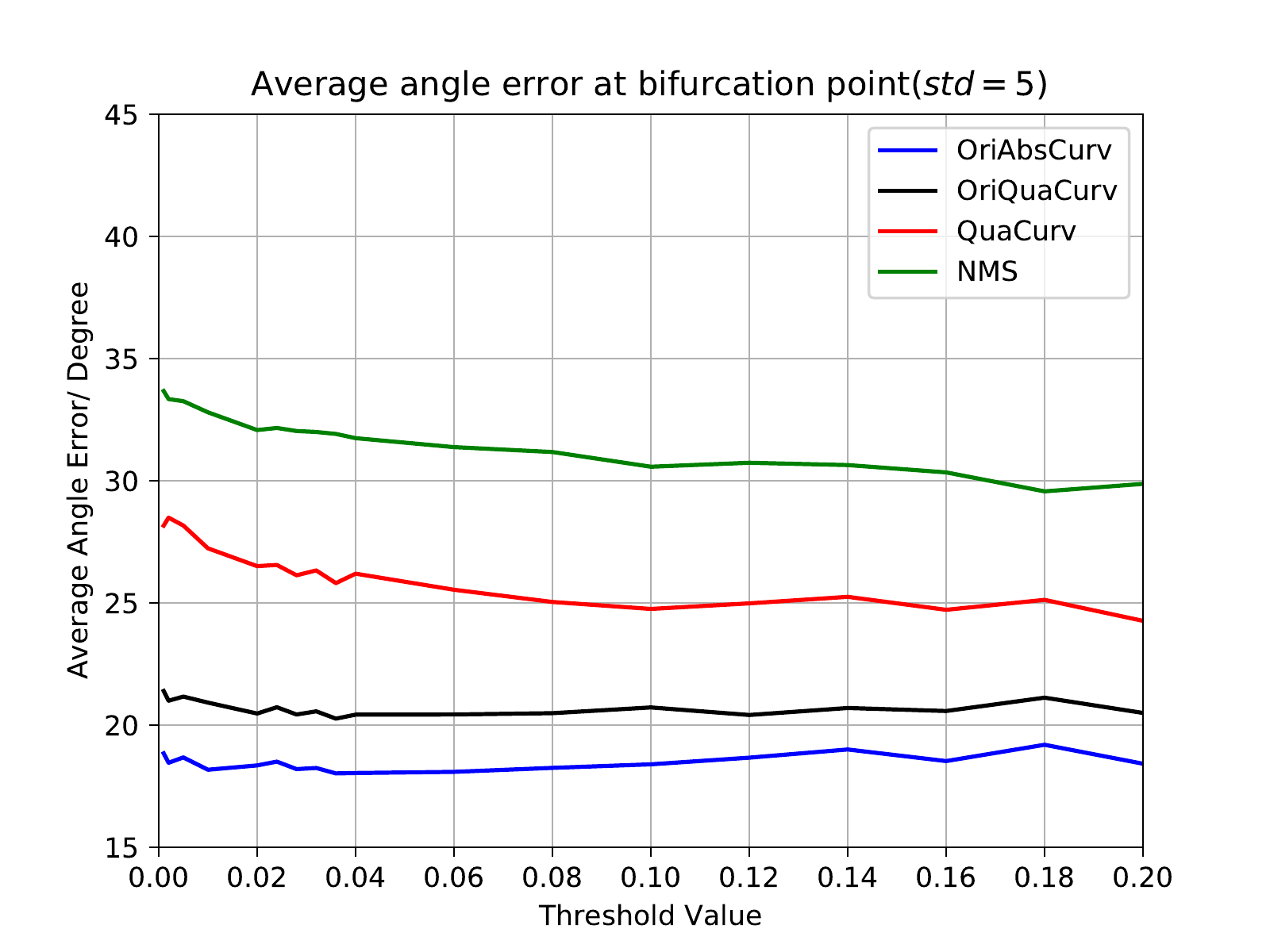}
\includegraphics[width=0.33\linewidth]{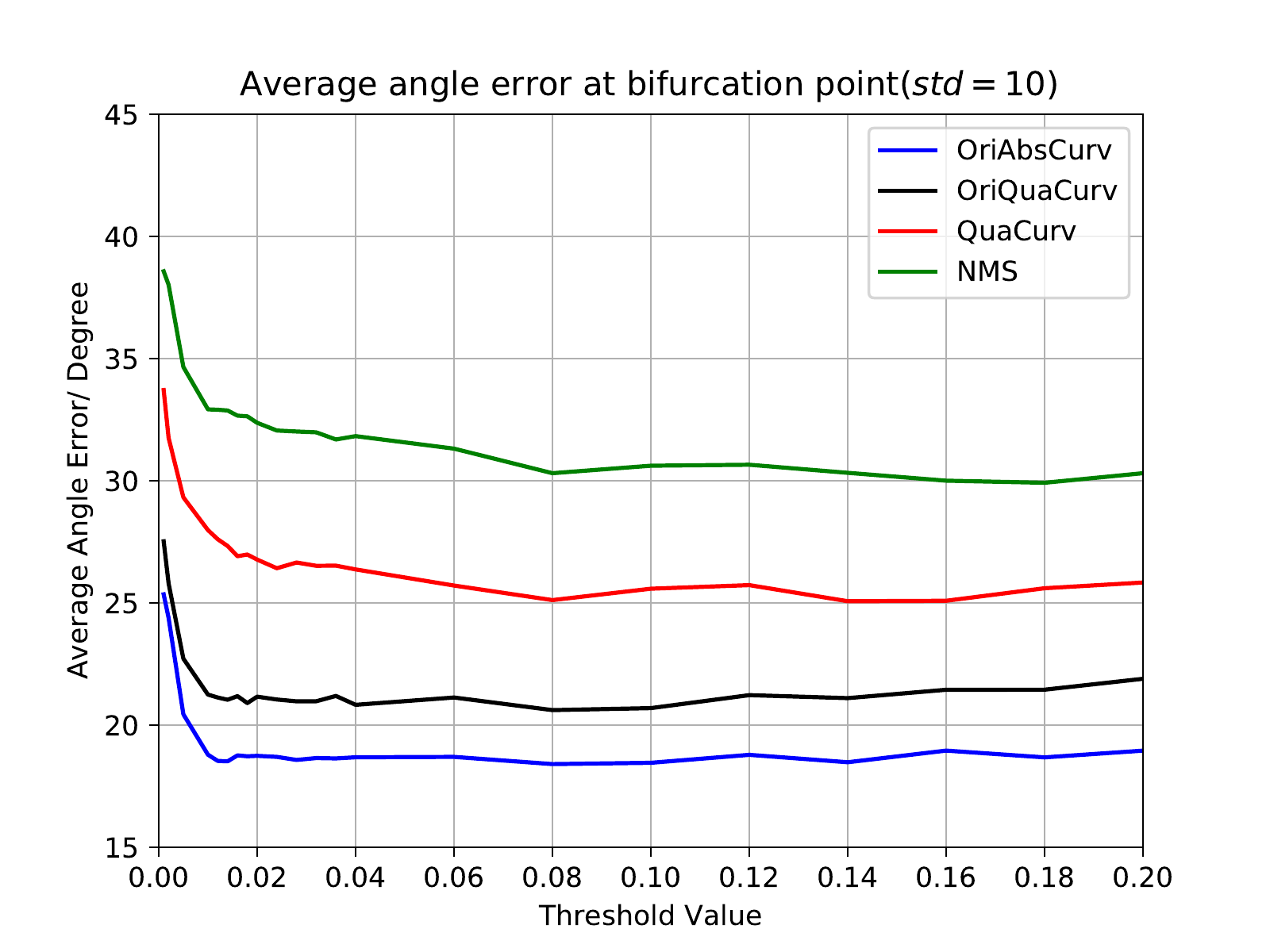}
\includegraphics[width=0.33\linewidth]{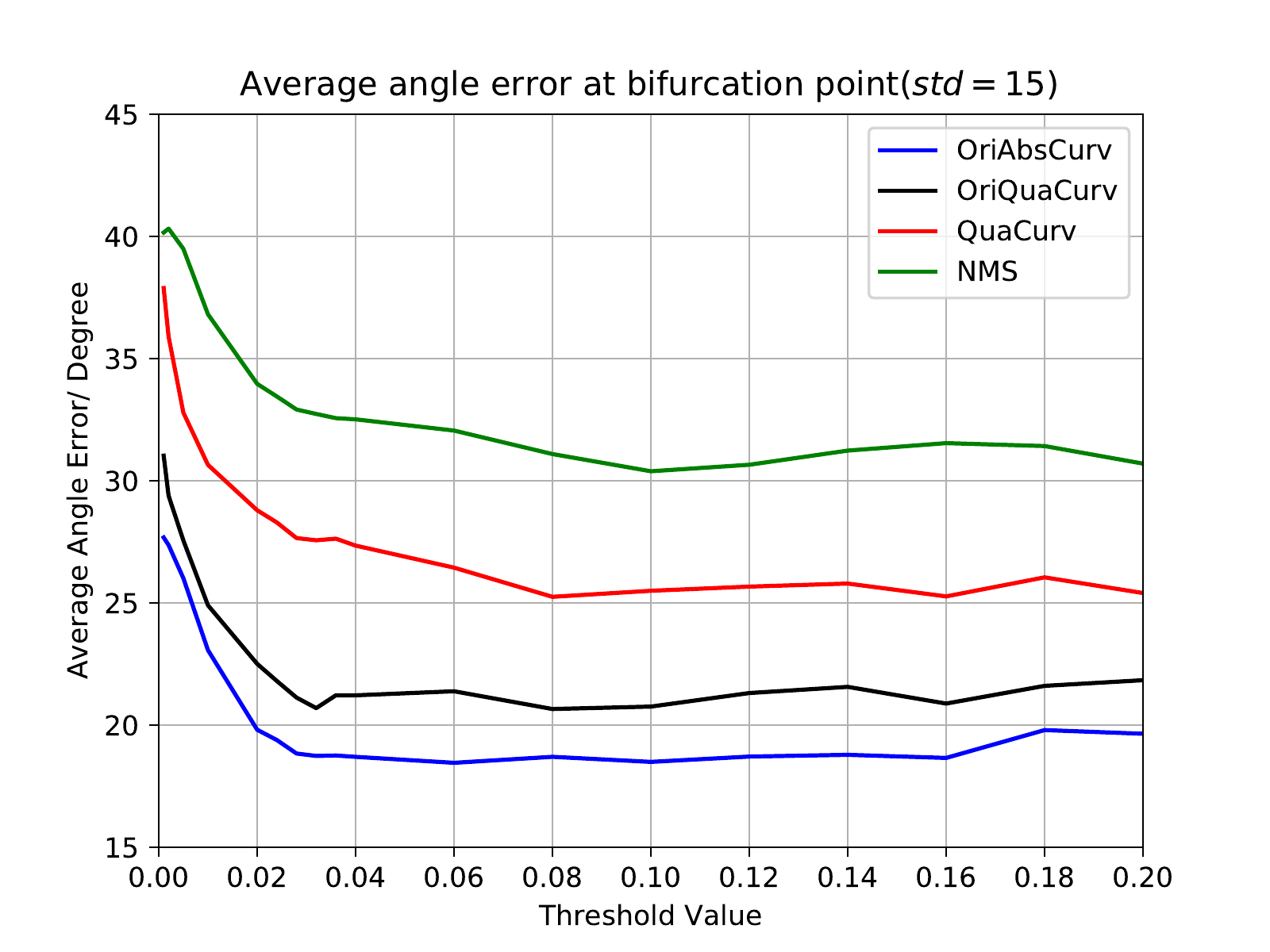}
\caption{Angle error comparison.}
\label{fig:angleError}
\end{figure*}

We used the modification\footnote{The implementation of \cite{hamarneh2010vascusynth} contains bugs, which were fixed.} of a method generating synthetic 3D vessel tree data \cite{hamarneh2010vascusynth}. The generated data consists of CT\footnote{Computer tomography}-like volume and ground truth vessel centerline tree, see Fig.~\ref{fig:wholeVolumeSynData} for an example. We generate 15 artificial volumes $100\!\times\!100\!\times\!100$ containing synthetic vascular trees with voxel intensities in the range $0$ to $512$. The size of voxel is $0.046$ mm. We use three different levels of additive Gaussian noise \cite{lehmann2010gaussiannoise} with standard deviations 5, 10 and 15.

\setlength\columnsep{3pt}

\noindent\textbf{Evaluation setup.} Our evaluation system follows \cite{thin:iccv15}. We first apply Frangi filter \cite{frangi1998multiscale} with hyperparameters $\alpha=0.5$, $\beta=0.5$, $\gamma=30$, $\sigma_{min}=0.023$ mm and $\sigma_{max}=0.1152$ mm. The filter computes {\em tubularness} \begin{wrapfigure}{r}{0.25\linewidth}
\vspace{-5mm}
\includegraphics[width=\linewidth]{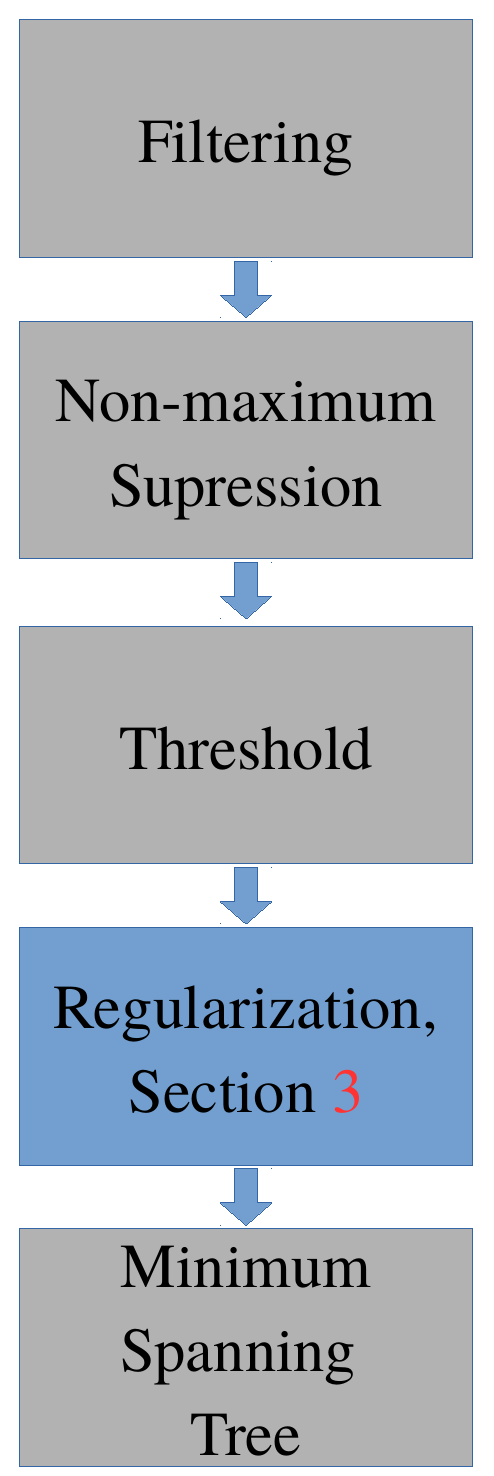}
\vspace{-10mm}
\end{wrapfigure}  {\em measure} and estimates tangent $l_p$ at each voxel $p$. Then we threshold the tubularness measure to remove background pixels. Then we use non-maximum suppression\footnote{The use of NMS is mainly for data reduction. Our method is able to work on thresholded data directly, see Fig. \ref{fig:triangleArtifact}(d).} (NMS)  resulting in voxel set $\Omega$. We use 26-connected neighborhood system $N$.  Next, we optimize our new join energy \eqref{eq:joint energy} to disambiguate tangent orientation and estimate centerline location, see Sec.~\ref{sec:joint_energy}. The hyperparameters are $\gamma=3.80$ (see energy \eqref{eq:oriented curv energy}), $\lambda=18.06$ (see energy \eqref{eq:joint energy}), $\tau=\cos70^{\circ}$ (see equation \eqref{eq:oricurvature}), and the maximum number of iterations is 1500 for both TRWS and Levenberg-Marquardt. Finally, we extract oriented vessel tree centerline as the minimum spanning tree of the complete graph.

Energy \eqref{eq:joint energy} assumes quadratic curvature term \eqref{eq:sqr curv}. 
However, it is to replace it with \eqref{eq:abs curv} to get 
an absolute curvature variant of our energy. 

We evaluate different regularization methods including energy \eqref{eq:unoriented curv energy} (QuaCurv), energy \eqref{eq:joint energy} with either quadratic curvature (OriQuaCurv) or  absolute curvature (OriAbsCurv) within the system outline above. We also compare to a tracing method \cite{aylward2002initialization} and medial axis \cite{bouix2005flux}. 

We adopt \emph{receiver operating characteristic} (ROC) curve methodology for 
evaluation of our methods and \cite{bouix2005flux}. We compute \emph{recall} and \emph{fall-out} statistics 
of an extracted vessel tree for different levels of the threshold.
The computed statistics define ROC curve. 

\begin{figure*}[t]
\centering
\footnotesize
\setlength\tabcolsep{1pt}
\begin{tabular}{cccc}
\includegraphics[width=0.23\linewidth,height=0.15\linewidth,trim=5cm 0 5cm 10cm,clip=true]{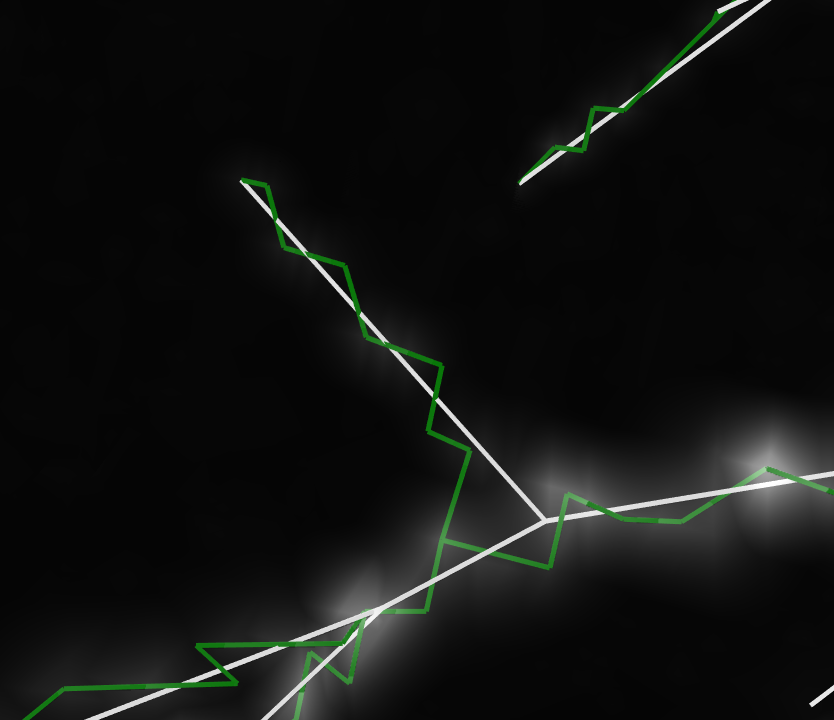}&
\includegraphics[width=0.23\linewidth,height=0.15\linewidth,trim=5cm 0 5cm 10cm,clip=true]{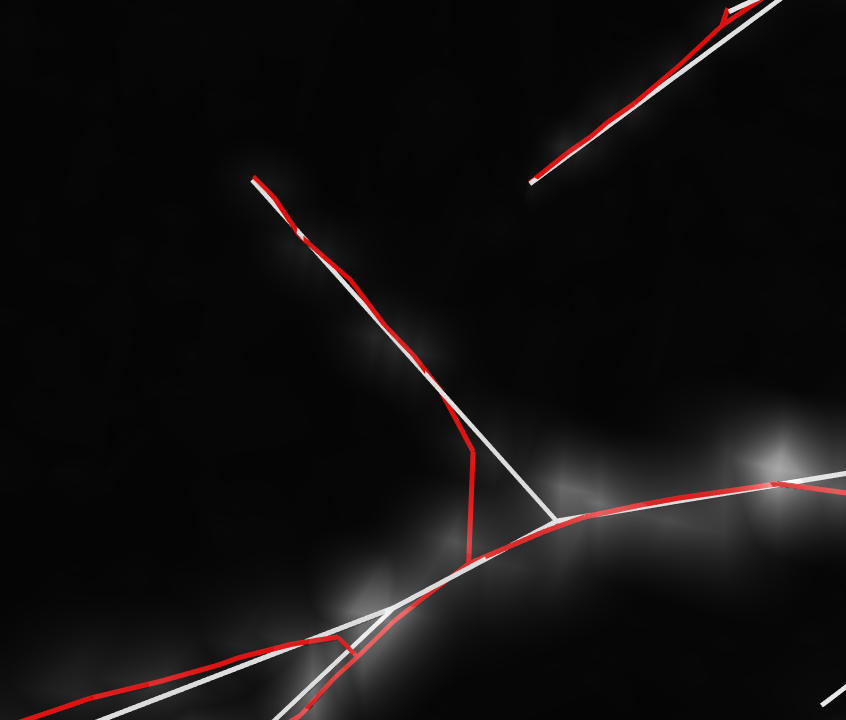} &
\includegraphics[width=0.23\linewidth,height=0.15\linewidth,trim=5cm 0 5cm 10cm,clip=true]{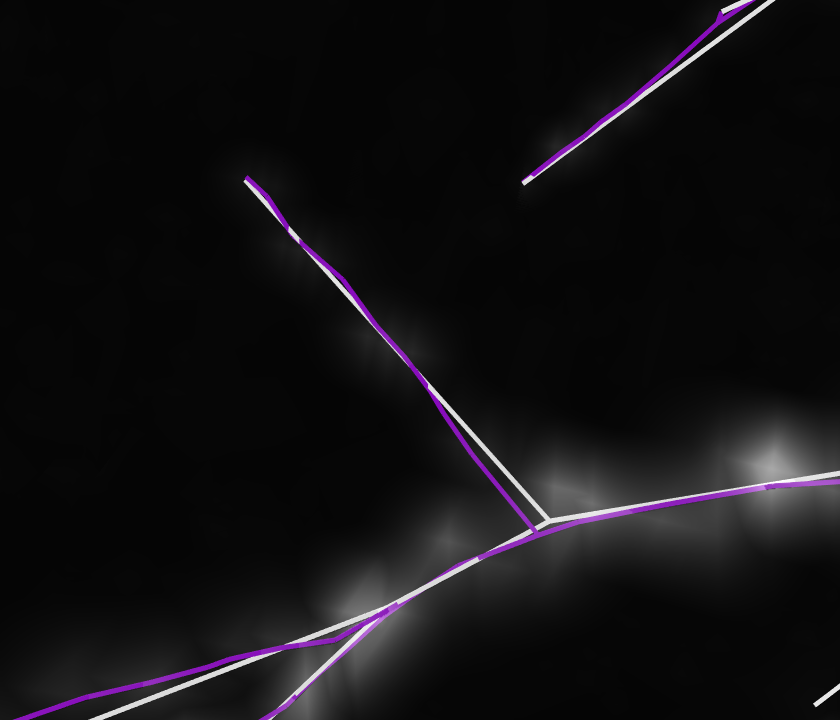} &
\includegraphics[width=0.23\linewidth,height=0.15\linewidth,trim=5cm 0 5cm 10cm,clip=true]{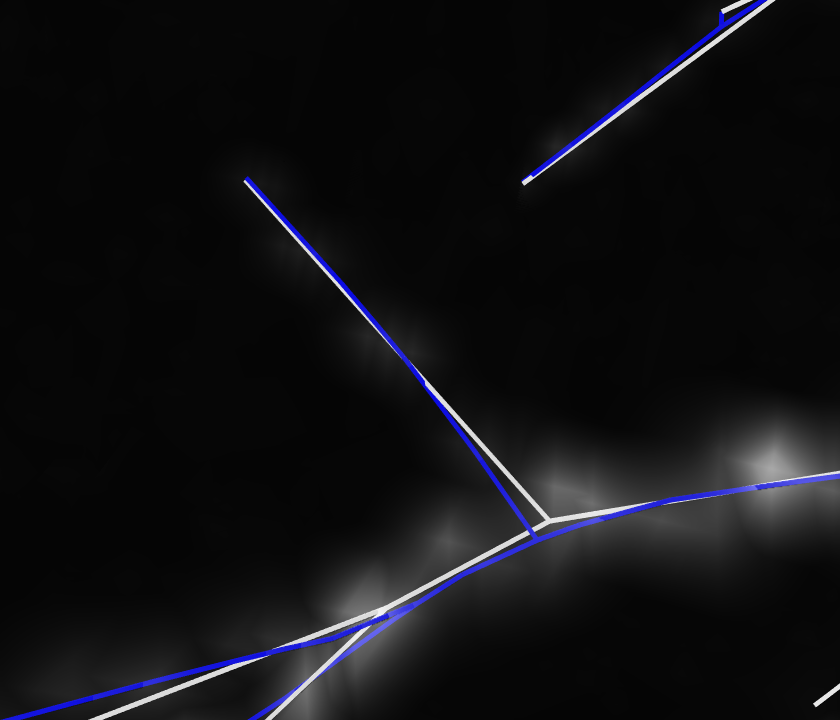}  \\ 
\includegraphics[width=0.23\linewidth,height=0.15\linewidth,trim=20cm 0 5cm 15cm,clip=true]{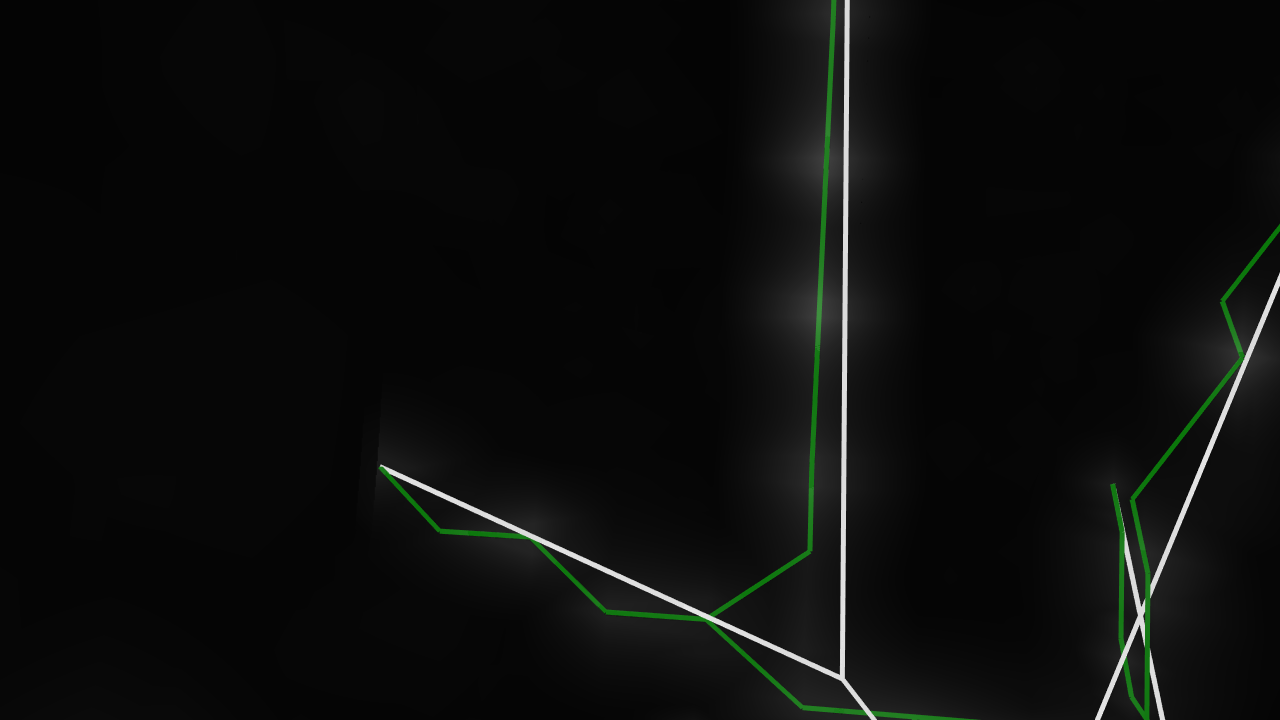} &
\includegraphics[width=0.23\linewidth,height=0.15\linewidth,trim=20cm 0 5cm 15cm,clip=true]{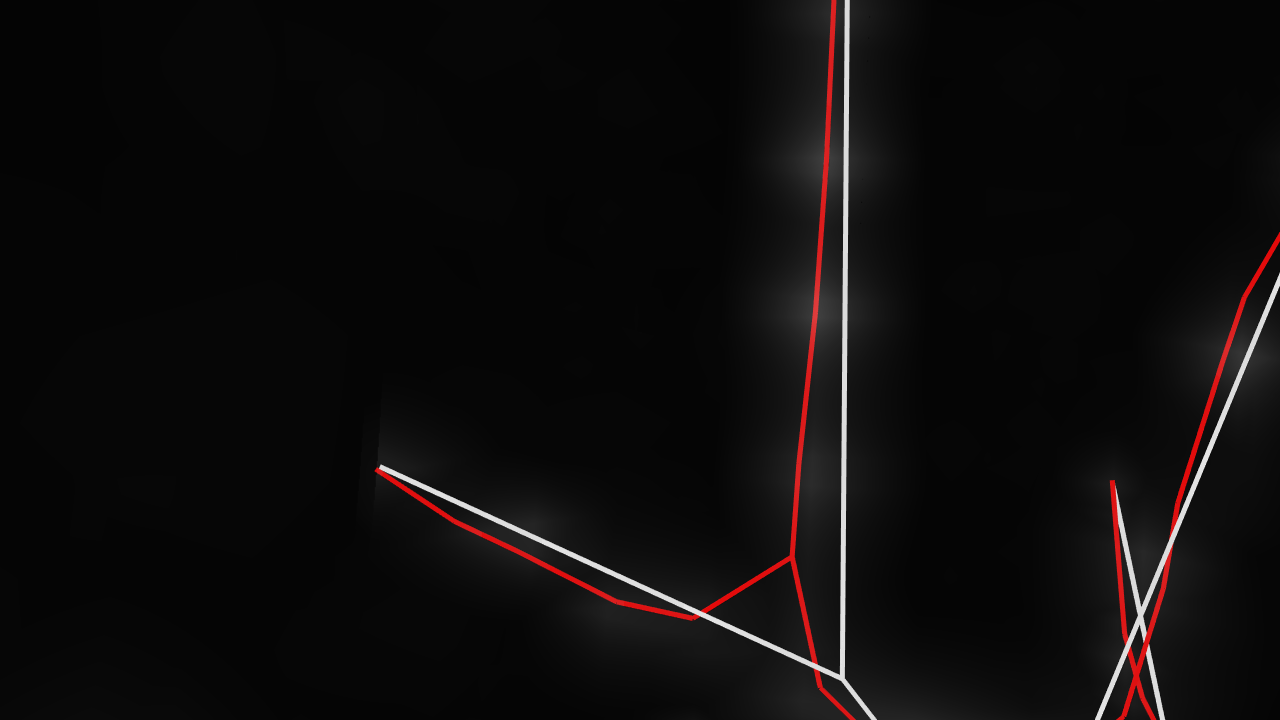} &
\includegraphics[width=0.23\linewidth,height=0.15\linewidth,trim=20cm 0 5cm 15cm,clip=true]{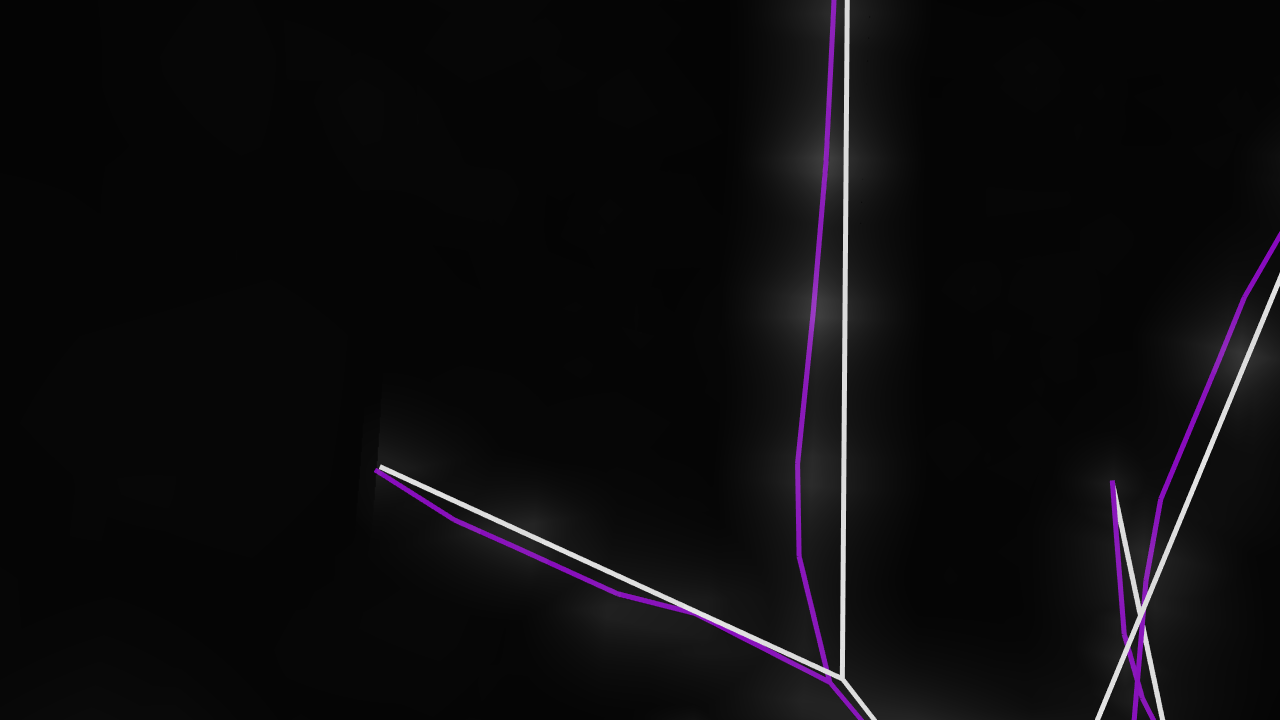} &
\includegraphics[width=0.23\linewidth,height=0.15\linewidth,trim=20cm 0 5cm 15cm,clip=true]{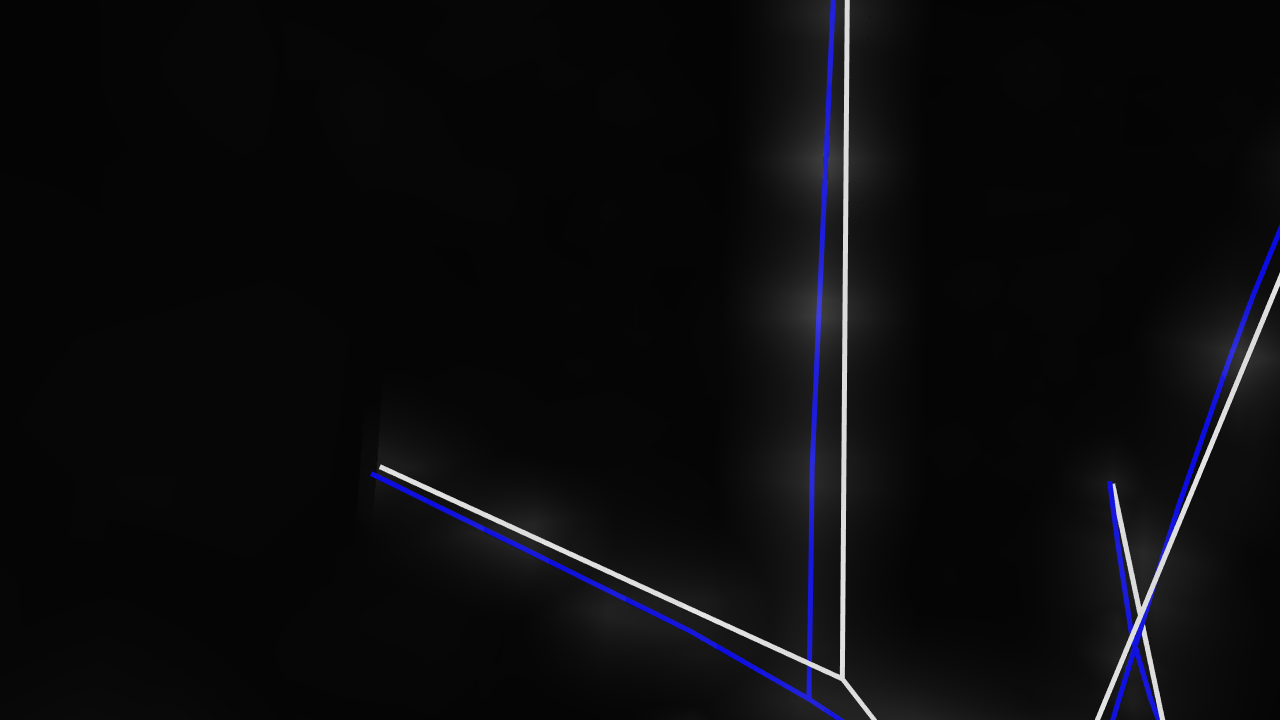} \\
\includegraphics[width=0.23\linewidth,height=0.15\linewidth,trim=20cm 2cm 8cm 12cm,clip=true]{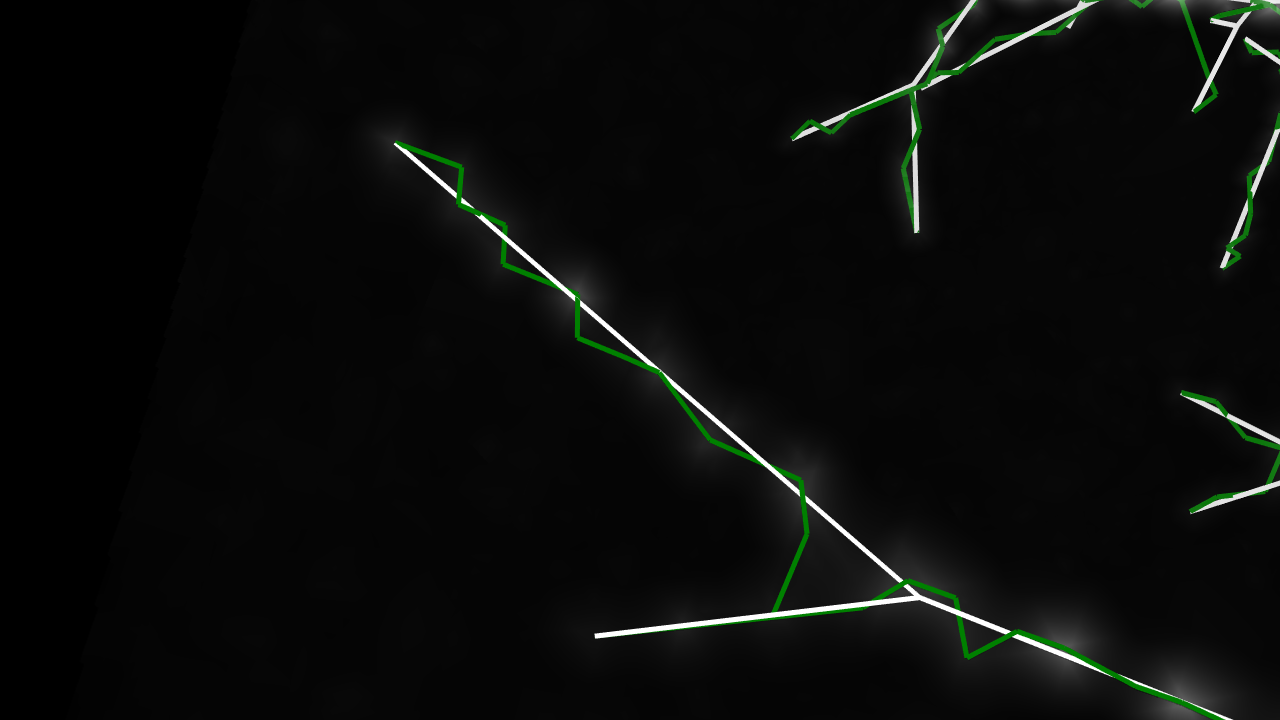} &
\includegraphics[width=0.23\linewidth,height=0.15\linewidth,trim=20cm 2cm 8cm 12cm,clip=true]{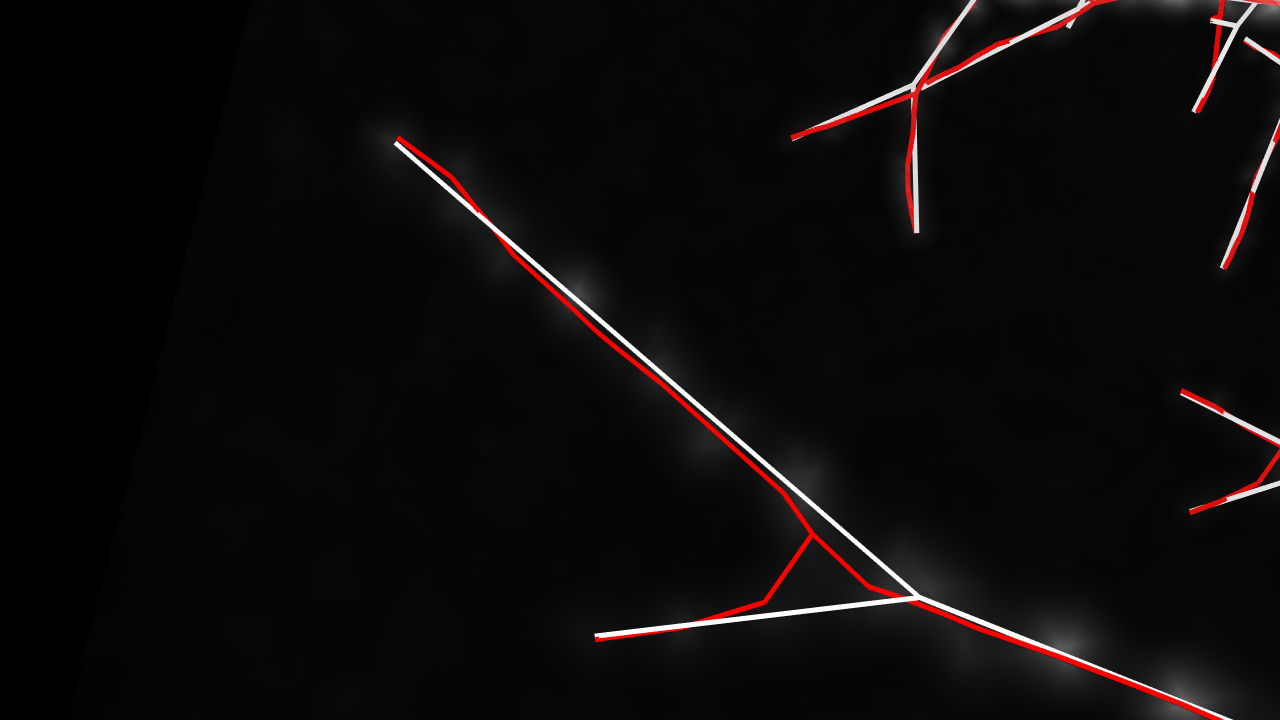} &
\includegraphics[width=0.23\linewidth,height=0.15\linewidth,trim=20cm 2cm 8cm 12cm,clip=true]{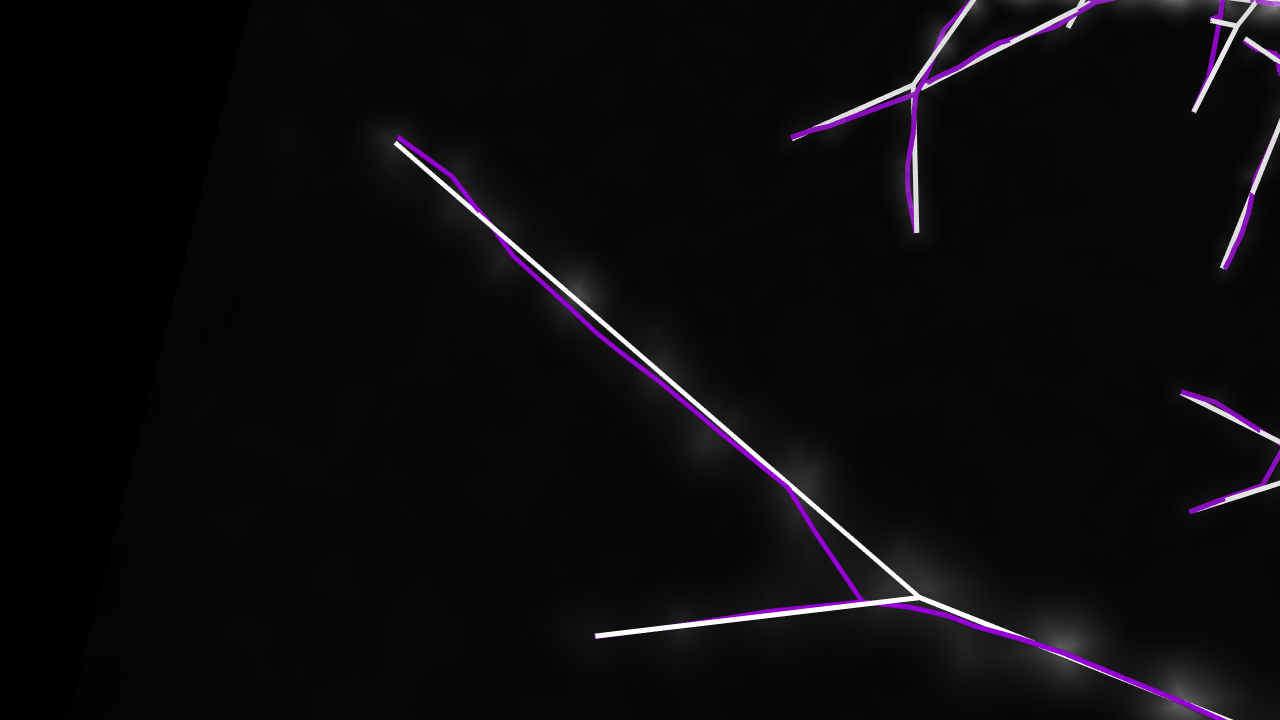} &
\includegraphics[width=0.23\linewidth,height=0.15\linewidth,trim=20cm 2cm 8cm 12cm,clip=true]{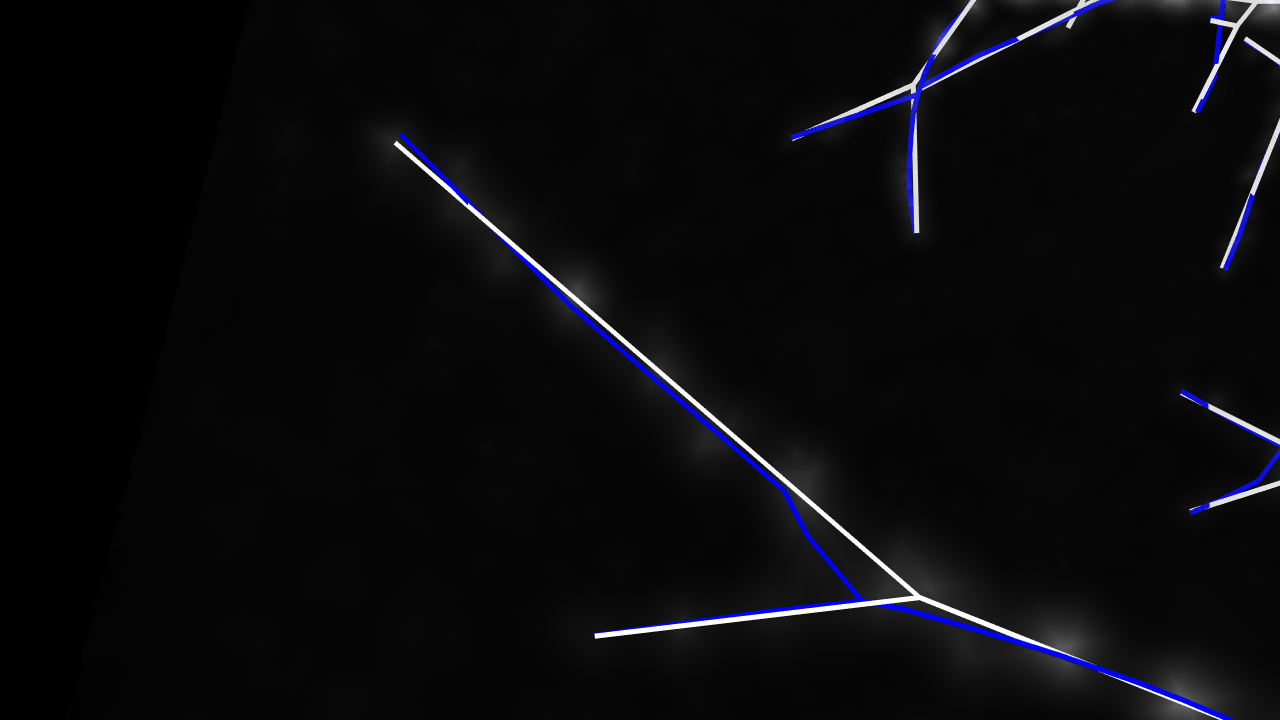} \\
(a)  & (b)  & (c) & (d)\end{tabular}
\caption{Examples of the result around bifurcations with regularization methods. White line is the ground truth tree. A tree extracted from NMS ouput directly (without regularization) is shown in (a). Solution of \eqref{eq:unoriented curv energy} \cite{thin:iccv15} is (b). Our model \eqref{eq:joint energy} is in (c). Our model \eqref{eq:joint energy} with absolute curvature is in (d).}
\label{fig:synExamples}
\end{figure*}

\begin{figure}[t]
\centering
\small
\includegraphics[width=0.95\linewidth,height=\linewidth]{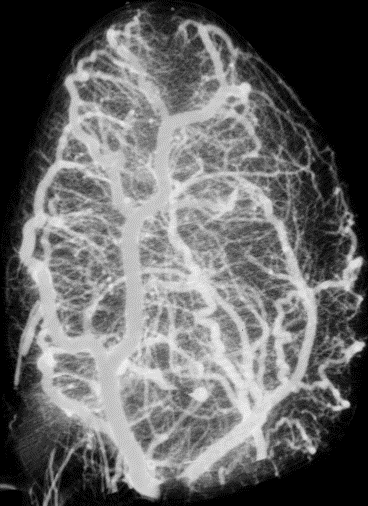}\\
(a) cardiac microscopy CT volume   \\
\includegraphics[width=0.95\linewidth,height=0.5\linewidth]{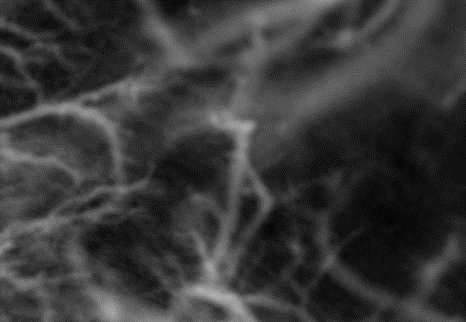}\\
(b) zoom-in
\caption{Visualization (MIP) of the raw volumetric data obtained from a mouse heart by {\em microscopic computer tomography}. The data is provided by Maria Drangova from the 
Robarts Research Institute in London, Canada.}\label{fig:realrawdata}%
\end{figure}

While ground truth is defined by locations at bifurcations and leaves of the tree, all evaluated methods yield densly sampled points on the tree. Therefore, we resample both ground truth and reconstructed tree with step size $0.0023$ mm. For each point on one tree, we find the nearest point on the other tree and compute the Euclidean distance. If the distance is less than $\max(r, c)$ voxels, this pair of points is considered a {\em match}. Here $r$ is the vessel radius at the corresponding point of the ground truth and $c=0.7$ is a matching threshold measured in voxels. The recall is 
$$\frac{N_{GTmatch}}{N_{GTtotal}}$$
where $N_{GTmatch}$ is the number of matched points in the ground truth and $N_{GTtotal}$ is the total number of points in the ground truth. The fall-out is 
$$1-\frac{N_{RTmatch}}{N_{RTtotal}}$$
where $N_{RTmatch}$ is the number of matched points in the ground truth  and $N_{RTtotal}$ is the total number of points in the ground truth. 

\begin{figure}[t]
\centering
\includegraphics[width=0.95\linewidth,height=\linewidth]{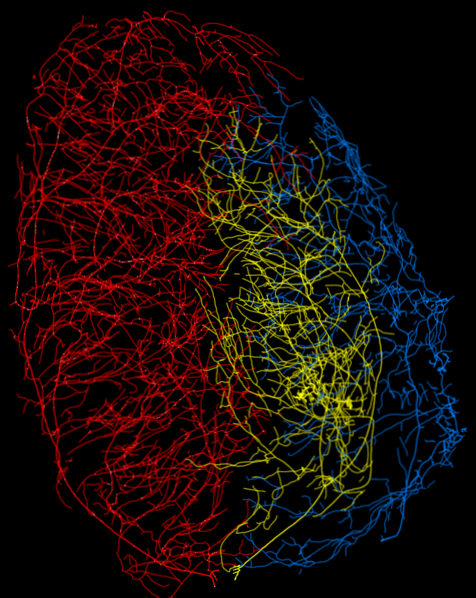}
\caption{Vessel tree reconstruction from real data in Fig.~\ref{fig:realrawdata}
based on our method for estimating centerline tangents using prior knowlegde 
about vessel divergence.
The final tree structure is extracted by MST on K-nearest-neighbour (KNN) weighted graph
with edge weights $w_{pq}$ defined as the average {\em arc-length} between 
neighbors $p$ and $q$ for two circles containing $p$ and $q$ and 
tangential to either $l_p$ or $l_q$. Three different colors (red, blue, yellow) 
denote three main branches.}
\label{fig:wholevolumerealresult}
\end{figure}

\begin{figure}[t]
\centering
\setlength\tabcolsep{1pt}
\begin{tabular}{cc}
\includegraphics[width=0.5\linewidth,height=0.35\linewidth]{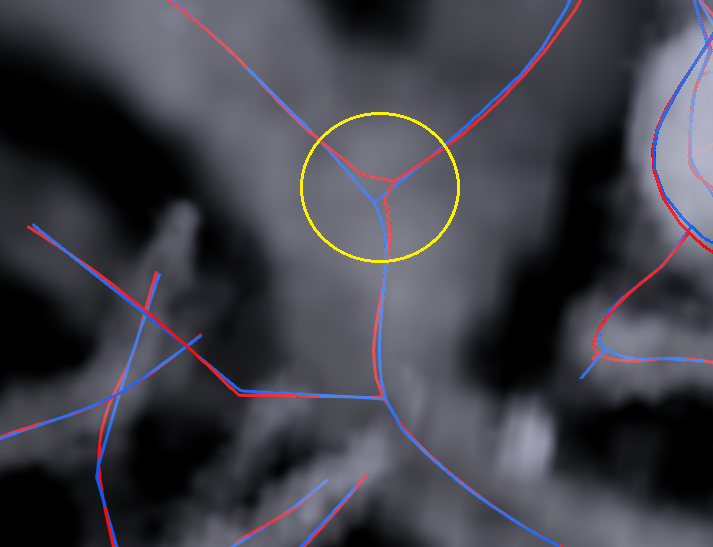}
&
\includegraphics[width=0.5\linewidth,height=0.35\linewidth]{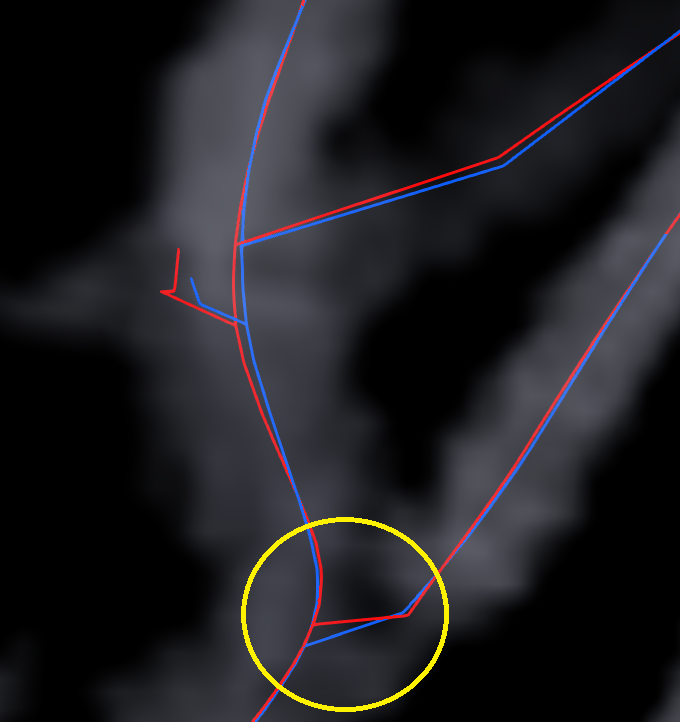}\\
\includegraphics[width=0.5\linewidth,height=0.35\linewidth]{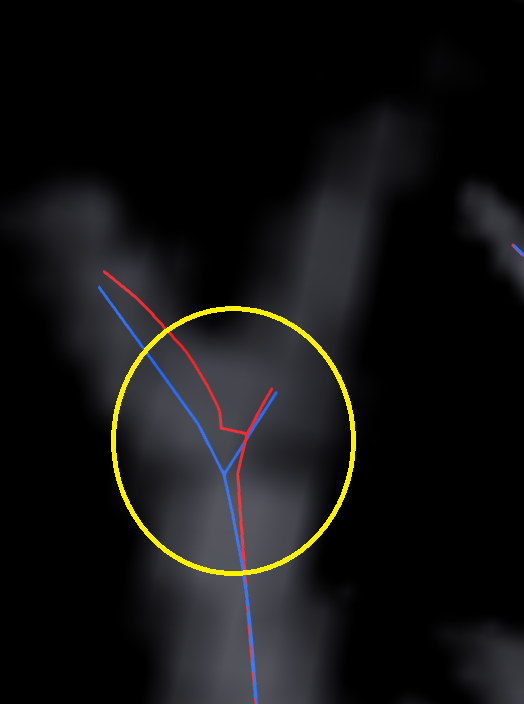}
&
\includegraphics[width=0.5\linewidth,height=0.35\linewidth]{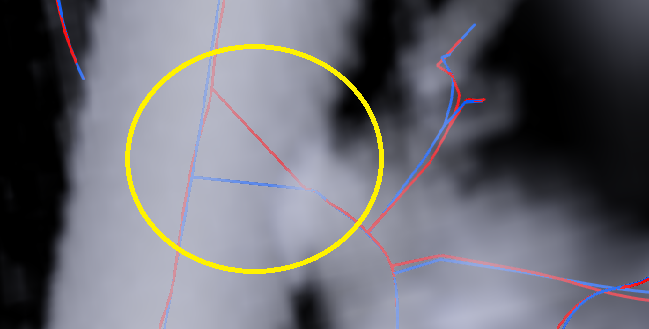}
\end{tabular}
\caption{Qualitative comparison results on real data. Red represents the result of \cite{thin:iccv15}, see \eqref{eq:unoriented curv energy}. Blue curve is the centerline obtained based on our 
directed vessel estimation model~\eqref{eq:joint energy} with divergence prior and absolute curvature regularization. The yellow circles highlight improvements at bifurcations due to correct estimation of the flow direction.}
\label{fig:realdatacomparison}
\end{figure}

The tracing method of \cite{aylward2002initialization} requires a seed points list as an input.
We generate four seed lists as described in Fig.~\ref{fig:rocResults}. The ROC curves in Fig.~\ref{fig:rocResults} favour our method. Since bifurcations is only a fraction of the data, the improvements around bifurcations are largely unnoticed in these  curves. Therefore, we compute the ROC curves for only bifurcation nodes. We use a bigger matching threshold $c=\sqrt{3}$ voxels. The results are shown in Fig.~\ref{fig:BifurRocResults} where the gap between methods is bigger. Also we compute angle errors at bifurcations, see Fig.~\ref{fig:angleError} and few examples in Fig.~\ref{fig:synExamples}.

\subsection{Real vessel data}
\label{sec:real}

We obtained the qualitative experimental results using a real micro-CT scan of
mouse's heart as shown in Figure \ref{fig:realrawdata}. The size of the volume
is $585\times 525\times 892$ voxels. Most of the vessels are thinner than voxel size. 
Due to the size of the volume the problem has higher computational cost than in Sec.~\ref{sec:synthetic}. We built custom GPU implementation of Levenberg-Marquardt algorithm to handle the large volume size. 
Figure~\ref{fig:wholevolumerealresult} shows the reconstructed centerline. 
Figure~\ref{fig:realdatacomparison} demostrate significant improvement of 
centerline estimation around bifurcations.

\section{Conclusions and Future work}
\label{sec:future}

We propose divergence prior for vector field reconstruction problems.
In the contest of vessel tree estimation, we use divergent vessel prior to estimate vessel directions
disambiguating orientations produced by Frangi filter. Our method significnatly improves the accuracy of reconstruction at bifurcations reducing the corresponding angle estimation errors by about 50 percent.

There are interesting extentions for our work on estimating vessel orientations. For example,
such orientations can be directly used for extracting vessel tree topology or connectivity. 
Instead of using standard MST on undirected graphs, \eg as in \cite{thin:iccv15}, 
we can now use Chu-Liu-Edmonds algorithm \cite{chu-liu:65,edmonds:67} to compute a minimum spanning 
{\em arborescence} (a.k.a. directed rooted tree)  on a directed weighted graph 
where a weight of any edge $(p,q)$ estimates the length of a possible direct ``vessel'' connection 
specifically from $p$ to $q$. Such a weight can estimate the {\em arc length} 
from $p$ to $q$ along a unique circle such that it contains $p$ and $q$, it is coplanar with $l_p$ and $q$,
and it is tangential to $l_p$. However, such constant curvature path from $p$ to $q$ works
as a good estimate for a plausible vessel connection from $p$ to $q$ only if $\langle l_p, pq \rangle >0$; otherwise there should be no edge from $p$ to $q$. This implies a directed graph since edges $(p,q)$ and $(q,p)$ will be
determined by two different tangents $l_p$ or $l_q$ and two different conditions $\langle l_p, pq \rangle >0$ or
$\langle l_q, qp \rangle >0$.  

\section*{Acknowledgements}
We would like to thank Maria Drangova (Robarts Research Institute, London, Ontario) for providing high-resolution microscopy CT volumes with cardiac vessels. We used TRWS code by Vladimir Kolmogorov (ITS, Vienna, Austria) for efficient minimization of binary orientation variables. Marc Moreno Maza (Western University, London, Ontario) shared his expertice in high-profirmance computing allowing our efficient implementation of trust region. This research would not be possible without support by the Canadian government including Discovery and RTI programs by NSERC.

%

\small
\bibliographystyle{ieee}   
\bibliography{ref}  

\end{document}